\documentclass[10pt,twocolumn,letterpaper]{article}

\usepackage[pagenumbers]{cvpr} 

\usepackage{graphicx}
\usepackage{amsmath}
\usepackage{amssymb}
\usepackage{mmstyle}
\usepackage{color}
\usepackage{booktabs}
\usepackage{algorithm}
\usepackage{diagbox}
\usepackage{multirow}
\usepackage{multicol}
\usepackage{longtable}
\usepackage{rotating}
\usepackage{multirow}
\usepackage[accsupp]{axessibility}

\newlength\savewidth\newcommand\shline{\noalign{\global\savewidth\arrayrulewidth
		\global\arrayrulewidth 1pt}\hline\noalign{\global\arrayrulewidth\savewidth}}

\usepackage[pagebackref,breaklinks,colorlinks]{hyperref}
\usepackage[capitalize]{cleveref}
\crefname{section}{Sec.}{Secs.}
\Crefname{section}{Section}{Sections}
\Crefname{table}{Table}{Tables}
\crefname{table}{Tab.}{Tabs.}
\def\cvprPaperID{501} 
\def\confName{CVPR}
\def\confYear{2022}

\begin{document}

\title{Towards Diverse and Natural Scene-aware 3D Human Motion Synthesis}

\author{
	Jingbo Wang$^1$
	~
	Yu Rong$^1$
	~
	Jingyuan Liu$^2$
	~
	Sijie Yan$^1$
	~
	Dahua Lin$^1$
	~
	Bo Dai$^3$ \\
	\normalsize
	$^1$ The Chinese University of Hong Kong 
	$^2$ Hong Kong University of Science and Technology \\
	\normalsize
	$^3$ S-Lab, Nanyang Technology University \\
	\normalsize
	$\{$wj020,ry017,dhlin$\}$@ie.cuhk.edu.hk, jliucb@connect.ust.hk, yysijie@gmail.com, bo.dai@ntu.edu.sg
}

\let\oldtwocolumn\twocolumn
\renewcommand\twocolumn[1][]{%
	\oldtwocolumn[{#1}{
		\begin{center}
			\includegraphics[width=\linewidth]{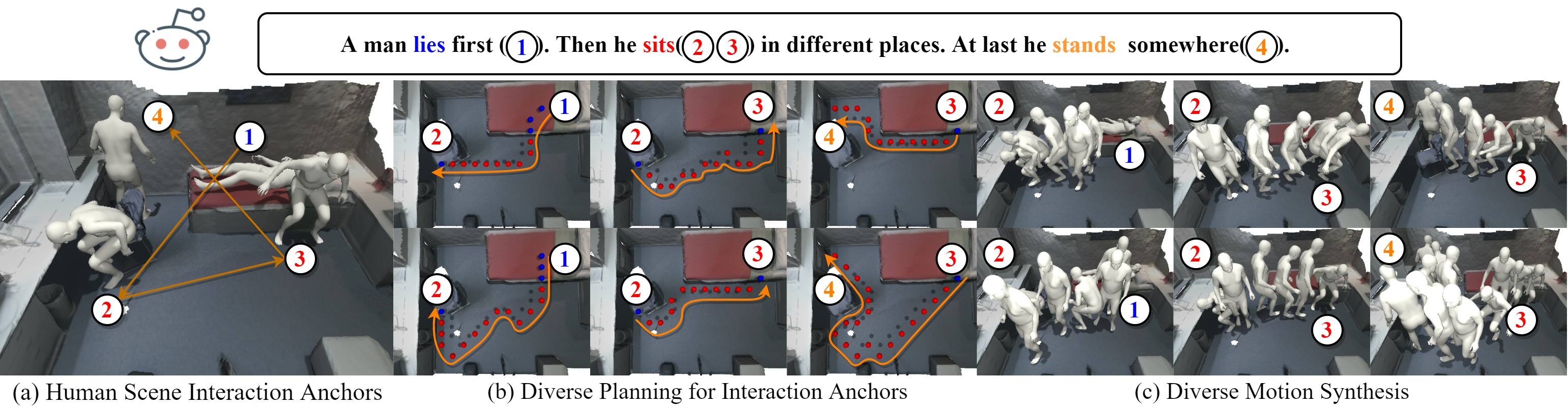}
			\captionof{figure}{\small
			We decompose scene-aware human motions into three aspects, namely human-scene interaction anchor, path planning, and body movements. Given the scene and the target action sequence, our framework first adopts specified schemes to generate diverse intermediate results for each aspect. These results are then integrated into miscellaneous yet coherent human motions.}		
	\label{fig:teaser}
\end{center}
}]
}

\maketitle

\begin{abstract}

The ability to synthesize long-term human motion sequences in real-world scenes can facilitate numerous applications.
Previous approaches for scene-aware motion synthesis are constrained by pre-defined target objects or positions and thus limit the diversity of human-scene interactions for synthesized motions. 
In this paper, we focus on the problem of synthesizing diverse scene-aware human motions under the guidance of target action sequences.
To achieve this, we first decompose the diversity of scene-aware human motions into three aspects, namely interaction diversity (\eg~sitting on different objects with different poses in the given scenes), path diversity (\eg~moving to the target locations following different paths), and the motion diversity (\eg~having various body movements during moving).
Based on this factorized scheme, a hierarchical framework is proposed, with each sub-module responsible for modeling one aspect.
We assess the effectiveness of our framework on two challenging datasets for scene-aware human motion synthesis.
The experiment results show that the proposed framework remarkably outperforms previous methods in terms of diversity and naturalness.

\end{abstract}

\section{Introduction}~\label{sec:intro}

The capability of synthesizing long human motion sequences is essential for a number of real-world applications, such as virtual reality and robotics.
Beyond early attempts that consider body movement synthesis in isolation~\cite{barsoum2018hp, cai2018deep, yang2018pose, yan2019convolutional,zhang2021we}, recent works~\cite{cao2020long,wang2021synthesizing, Wang_2021_CVPR, hassan_samp_2021} begin to explore the influences of surrounding scenes on human motion synthesis for different actions.
Limited by the 2D representation of scene context~\cite{cao2020long,Wang_2021_CVPR} or the reliance on manually assigned interacting targets~\cite{wang2021synthesizing,hassan_samp_2021},
these approaches mainly focus on modeling the body movements and fail to comprehensively investigate the inherent diversity of scene-aware human motions.
In order to synthesize long-term human motions guided by the scene context and the target action sequence,
we propose to model the inherent motion diversity across different granularities, each contributing to different aspects of human motion.

As shown in Figure~\ref{fig:teaser}, the diversity of scene-aware human motions can be factorized into three levels, given the target action sequence~(\eg~A man lies first. Then he sits in different places. At last, he stands somewhere.).
Firstly, given the surrounding scene context and the target action sequence, there exists a distribution of valid locations to realize the actual human-scene interactions for each of these actions~(\eg~We can sit on any chairs or beds and stand on the ground).
Different locations can be sampled from the distribution and serve as the anchors of the whole synthesized motion sequence.
Based on those anchors, we can then follow various paths to bridge them one by one.
Finally, our body poses also differ from case to case when we move along the paths to connect all anchors.
We demonstrate these three levels of diversity in Figure~\ref{fig:teaser}.
Existing attempts for scene-aware human motion synthesis~\cite{wang2021synthesizing,hassan_samp_2021} only emphasize the last level of diversity~(\eg~walking to the pre-defined object or position in the scene) via manually assigning the interaction locations and motion paths.
Consequently, the importance of the scene semantics is substantially muted, as it mainly affects the distribution of valid interaction anchors and the distribution of valid motion paths.
To faithfully capture the diversity of scene-aware human motions, we propose a novel three-stage motion synthesis framework, each stage of which is responsible for modeling one level of the aforementioned diversity.

For \textbf{diverse human-scene interaction anchors}, we design our pose placing framework for the given action sequence.
Different from~\cite{PSI:2019, PLACE:3DV:2020} which only consider the influence of scene context, we first synthesize scene-agnostic poses according to the target action via a conditional VAE (CVAE)~\cite{sohn2015learning}.
Then we follow the practice of POSA~\cite{Hassan:CVPR:2021} to place these poses into the scene.
To be specific, the 3D scene is uniformly split into a set of non-overlapping grids, each of which is associated with a validity score that measures its compatibility as a candidate for placing the poses.
We make two modifications to the original placing method used by POSA
First, we introduce the position relationship between poses with the same action label to enhance the placing diversity by avoiding them being placed to the nearby positions.
Furthermore, we leverage another CVAE model as the placing refiner to produce diverse offsets for each discrete grid.
Examples of generated anchors are depicted in Figure~\ref{fig:teaser} (a).

To produce \textbf{diverse obstacle-free motion paths} following the sampled anchors, we employ an adapted A$^*$ algorithm over the discrete 3D grids as the path planner.
The standard A$^*$ algorithm used by previous works~\cite{hassan_samp_2021} only generates deterministic paths as they only consider collision between objects and distances to the target locations.
To model the inherent diversity of motion paths, we amend the original algorithm with a trainable stochastic module learned in a data-driven manner.
The new module, named Neural Mapper, can provide dynamic scene-conditioned probabilistic guidance to the A$^*$ algorithm, so that the algorithm can automatically produce diverse yet natural paths given the deterministic scenes and location anchors. 
We show several examples of generated diverse paths given the same start and end locations in Figure~\ref{fig:teaser} (b).

Lastly, we propose a novel Transformer-based CVAE, called motion completion network, to synthesize \textbf{diverse body movements} guided by the paths generated in the previous step.
Inspired by~\cite{petrovich21actor}, we leverage Transformer as the basic architecture for synthesizing continuous and smooth motions.
Differently, we focus on diverse motion completion of poses with long-term distance and different actions, rather than synthesize motions for the single action~\cite{petrovich21actor}.
Therefore, this motion completion network first generates diverse moving trajectories, loosely following the paths sampled by the aforementioned A$^*$ algorithm.
The body poses are then produced by taking the scene contexts, action labels, human-scene interaction anchors, and synthesized trajectories as inputs.

To summarize our contributions:
1) We analyze the \textbf{inherent diversity} of the human motion and decompose it into three components, namely the diversity on human-scene interaction anchors, paths, and body poses.
2) We propose a novel three-stage framework to \textbf{faithfully capture the diversities} of scene-aware human motions.
This framework can automatically synthesize human motions following these diversities with the condition action labels.
Qualitative and quantitative results on datasets such as PROX~\cite{PROX:2019}demonstrate that our method significantly surpasses previous approaches in terms of diversity and naturalness.
3) In the proposed framework, we make several technique contributions for this task, including the action conditioned pose placing framework for generating diverse human-scene interaction anchors, Neural Mapper for planning diverse paths, and motion completion network for producing diverse and continuous motions. With our decomposition on motion diversity, these technique contributions can achieve our goal efficiently and effectively.

\section{Related Works}
\begin{figure*}[t]
	\centering
	\includegraphics[width=0.9\textwidth]{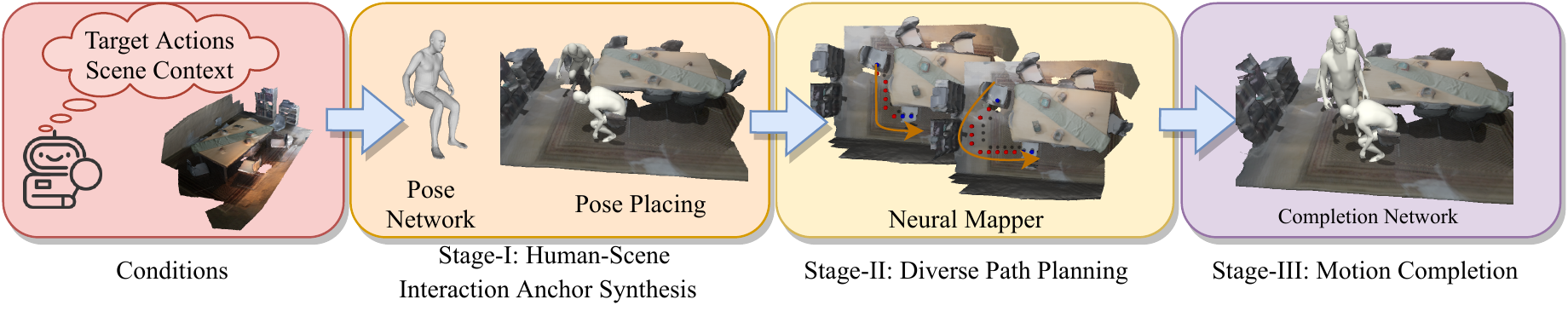}
	\caption{\small\textbf{Overview of the framework.} Our framework is composed of three stages. Given the target actions and the scene contexts, our framework first generates human-scene interaction anchors by firstly synthesizing scene-agnostic poses via the pose network and then placing the poses into the scene guided by the scene contexts and Pose Refiner. Then the framework produces diverse planning paths through the adapted A$*$ algorithm that is amended with the novel Neural Mapper. At last, the motion completion network is leveraged to synthesize natural human motions guided by the anchors and following the planned paths.}
	\label{fig:framework}
\end{figure*}

\paragraph{Motion Synthesis.}
Early works~\cite{barsoum2018hp, cai2018deep, yang2018pose,yan2019convolutional,harvey2020robust, xu2020hierarchical, pavllo2018quaternet} focus on synthesizing natural body poses and neglect the influences of other factors such as action and environments.
Recent studies begin to explore the relationship between human motions with actions and scene contexts.
Recent works~\cite{chuan2020action2motion,cai2021unified} generates human pose sequences with a CVAE model~\cite{sohn2015learning} based on the given action labels.
ACTOR~\cite{petrovich21actor} builds up a transformer based on CAVE to synthesize human motion sequence directly from the given action label.
Cao~\emph{et al.}~\cite{cao2020long} propose a three-stage motion prediction method that can predict different human motions with different destinations.
Wang~\emph{et al.}~\cite{Wang_2021_CVPR} extend CSGN~\cite{yan2019convolutional} to explore the influence of 2D scene contexts on human motion synthesis.
Wang~\emph{et al.}~\cite{wang2021synthesizing} build up a framework to synthesize human motions in the 3D scene controlled by the given pairs of begin-end points.
SAMP~\cite{hassan_samp_2021} extends~\cite{StarkeZKS19} to use 3D oriented objects to facilitate the synthesis of human motions with specific action labels.
Besides, a planning module is incorporated into their framework to find obstacle-free paths.

The limitations of previous works~\cite{hassan_samp_2021,wang2021synthesizing} mainly lie in their reliance on predefined objects or positions, which constrain their ability to explore the inherent interaction diversity of synthesized scene-aware human motions.
In this work, we aim to overcome the limitations of the previous works and synthesize diverse motions guided by target action sequences in the given scenes.
To achieve this, we first synthesize diverse human-interaction anchors, which interacts with different objects in the scene.
Then we plan diverse paths and complete diverse body movements between these anchors.

\paragraph{Motion Prediction.}
Motion prediction is closely related to our problem.
Different from the motion synthesis, the goal of this task is to predict human dynamics in the future with the given moving orientations or previous motions.  
Martinez~\emph{et al.}~\cite{martinez2017human} and ERD~\cite{fragkiadaki2015recurrent} proposed motion prediction framework based on the Seq2Seq model~\cite{sutskever2014sequence}.
Ac-LSTM~\cite{li2017auto} mixes synthesized frames and observed frames to enhance the capability of LSTM~\cite{hochreiter1997long} during the training stage.
The graph convolution network~\cite{KipfW17, yan2018spatial}is widely used in recent motion prediction~\cite{Li_2020_CVPR, Cui_2020_CVPR, mao2019learning}.
These methods model dynamic spatial and temporal relationships between the obvious frames and the future frames. 
Different from these works, our goal is to synthesize motions without prior knowledge of the previous motions.
\section{Methodology}
\subsection{Overview}
We first formally define the task of scene-aware 3D human motion synthesis.
We use triangular mesh $S = (v^s, f^s)$ to represent the scene context, where $v^s$ and $f^s$ stand for vertices and faces.
Our task is to synthesize diverse 3D human motions in the given scene context $S$, driven by a sequence of target action labels $A = (a_1, a_2, ..., a_N)$.
Each label stands for one scene-related human action, such as sitting or laying.
The synthesized 3D human motions are represented as a sequence of SMPL-X models~\cite{pavlakos2019expressive} described by their parameters $\{P_0, ..., P_T\}$,
where $P_i$ is composed of $(t_i, \phi_i, \theta_i)$,
$t_i \in \mathbb{R}^3$ is the global translation, $\phi_i \in \mathbb{R}^6$ is the global orientation represented in 6D continuous rotation~\cite{zhou2019continuity}.
$\theta_i \in \mathbb{R}^{32}$ is the body pose parameters, represented in the form of VPoser~\cite{pavlakos2019expressive}.
We use mean values for remaining SMPL-X parameters, including shape parameters, facial parameters, and hand poses.

The overview of our framework is depicted in Figure~\ref{fig:framework}.
We aim to solve this challenging problem in a hierarchical manner via exploiting the inherent properties of the scene-aware human motions.
Our framework first generates diverse human-scene interaction anchors for the given actions.
In this step, the framework first produces scene-agnostic poses corresponding to the action labels and then places these poses into the scene considering the compatibility between the synthesized poses and the scene.
In the next step, we leverage a path planning module to produce diverse obstacle-free paths under the guidance of the synthesized anchors from the first step.
Finally, a motion completion module is adopted to synthesize diverse body movements that fill in the missing motions between consecutive anchors while roughly following the planned paths from the second step.
In the following, we introduce our modules in detail. 

\subsection{Human-Scene Interaction Anchor Synthesis}
\label{subsec:interaction}

We first synthesize human-scene interaction anchors.
Unlike previous works~\cite{PSI:2019,PLACE:3DV:2020} that only condition human motion synthesis on the scene context,
we use action labels describing interaction types as an additional condition.
To be specific, we first synthesize scene-agnostic poses corresponding to the action labels.
Then we follow the practice of POSA~\cite{Hassan:CVPR:2021} with several modifications to diversely place the synthesized poses into the scene.
This design affords us more control over the final synthesized motions.

\begin{figure}[t]
	\centering
	\includegraphics[width=0.48\textwidth]{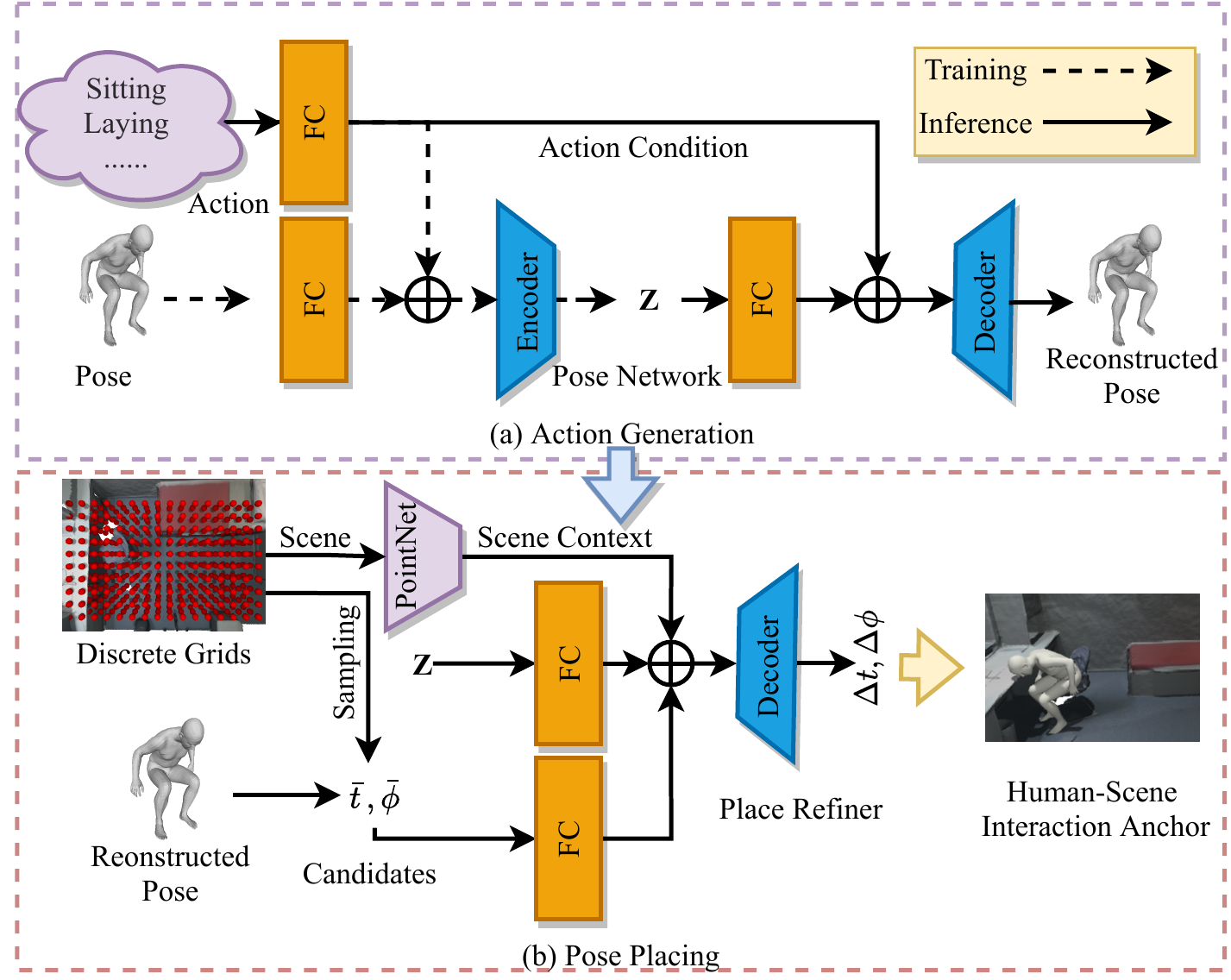}
	\small\caption{\textbf{Human-scene interaction anchor generation.} There are two steps to synthesize human-scene interaction anchors. The first step is generating diverse scene-agnostic poses conditioned on the target action, as shown in~(a). The second step is placing synthesized poses in the given scene, as depicted in~(b).
	}
	\label{fig:placing}
\end{figure}

\paragraph{Scene-Agnostic Pose Synthesis.}
As shown in Figure~\ref{fig:placing}~(a), we follow the standard CVAE framework to synthesize scene-agnostic poses $\theta_i$ with the target action $a_i$.
To be specific, we first sample noises from the prior Gaussian distribution and encode them with a fully-connected layer.
Then, we use the one-hot vector $a_i$ to represent the action condition and encode it with another fully-connected layer.
These two features are added up and then served as additional input besides the noise.
The model outputs the synthesized pose $\theta_i$, which is directly used as the body pose for the anchor $P_i$ for $i$-th anchor.

\paragraph{Scene-Conditioned Anchor Placing.}
In this step, we place the scene-agnostic poses $(\theta_1, \theta_2, ..., \theta_N)$ into the given scene.
There are two aspects to be taken into consideration in this step.
The first one is how to place poses to locations with compatible scene structure and interaction semantics.
The other one is how to efficiently find multiple reasonable locations given a pose.

Therefore, we first select our placing candidates following the practice of POSA~\cite{Hassan:CVPR:2021}.
Specifically, each candidate consists of a translation parameter $\bar t_i$ and an orientation parameter $\bar o_i$ for the anchor $P_i$.
We split the given scene into uniform non-overlapping discrete girds as translation candidates.
For each discrete gird, we then uniformly sample eight different orientations that are parallel with the ground plane to build orientation candidates.
Each translation candidate is paired with one of its associated orientations to form one placing candidate.
For each scene-agnostic pose $\theta_i$, we then rank all the placing candidates by their compatibility scores with the pose, which is proposed by~\cite{Hassan:CVPR:2021} that considers both the affordance and penetration.
An intuitive idea is to select the candidate with the best score.
However, our empirical study shows that candidates with the same action labels tend to be located close to each other since the same action usually shares similar physical and semantic structures, as shown in the first row of Figure~\ref{fig:placing_show}.
To increase the placing diversity, we introduce an additional penalty on the locations that have been occupied by anchors with the same action labels.
As shown in Figure~\ref{fig:placing_show}, this new penalty helps produce more diverse placing candidates for similar poses.
In this way, we can sample an initial placing candidate $(\bar t_i, \bar \phi_i)$ for each pose $\theta_i$.
The initial anchor $\bar P_i = (\bar t_i, \bar \phi_i, \theta_i) $ is then constructed subsequently.

In practice, we further adopt another sub-module called Place Refiner to improve the micro diversity of the placing candidates.
Place Refiner is implemented as a CVAE model that takes the noise of $\theta_i$, the scene context encoded by the PointNet~\cite{qi2017pointnet} and the initial anchor $\bar P_i$ as the input.
It outputs the offset $(\Delta t_i, \Delta \phi_i)$ to the sampled position and orientation $(\bar t_i, \bar \phi_i)$.
The final position and orientation are obtained as $t_i = \bar t_i + \Delta t_i$ and $\phi_i = \bar \phi_i + \Delta \phi_i$.
The framework of Place Refiner is depicted in Figure~\ref{fig:placing} (b).

\begin{figure}[t]
	\centering
	\includegraphics[width=0.48\textwidth]{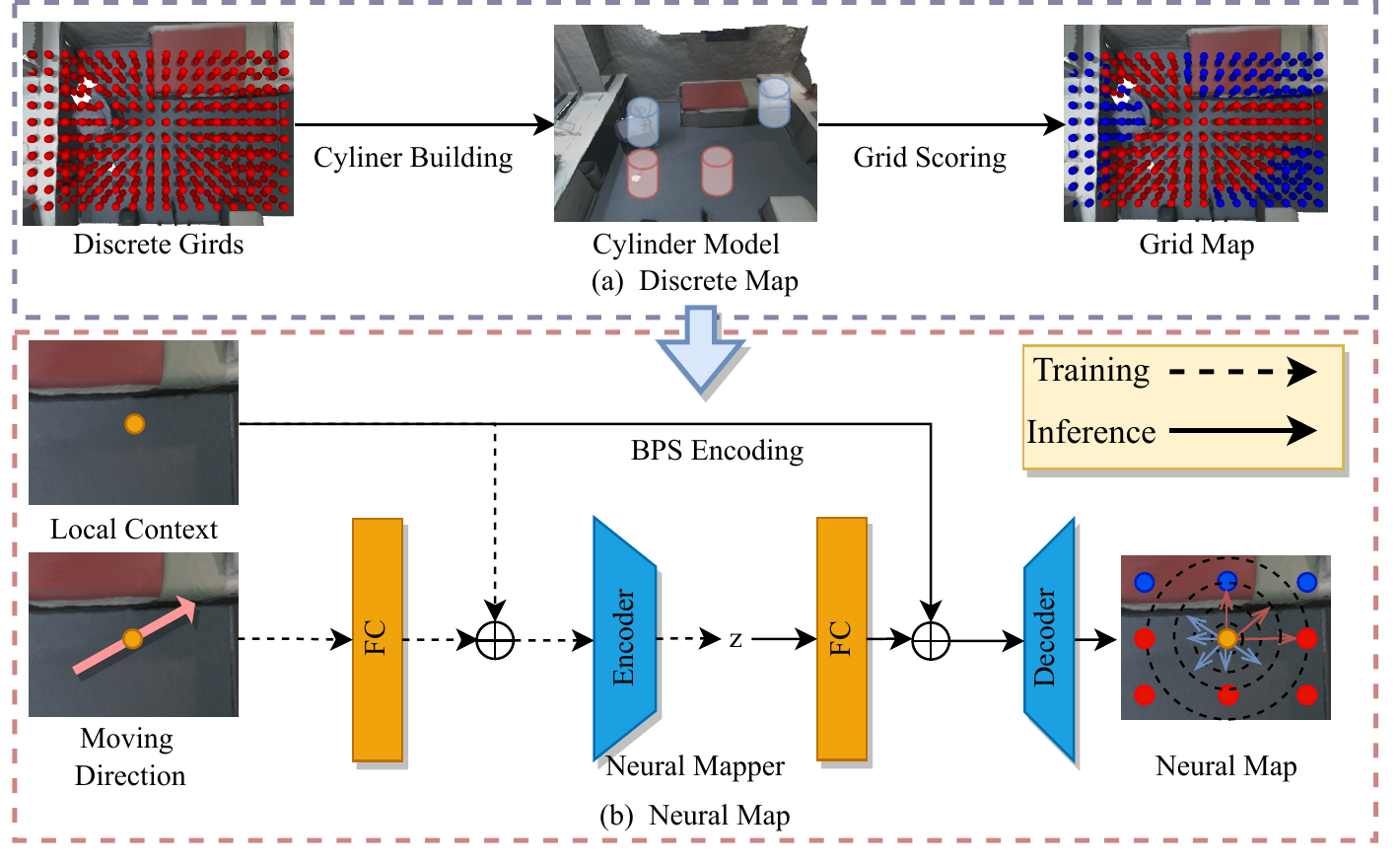}
	\small\caption{\textbf{Map building.} The map for our planning algorithm is built based on the collision detection (a) and Neural Mapper (b). Neural Mapper provides diverse moving probability for each neighbor gird to plan diverse paths.}
	\label{fig:map}
\end{figure}

\subsection{Diverse Path Planning}~\label{subsec:planning}

In this step, we discuss how to generate diverse obstacle-free paths from human-scene interaction anchors.
Previous works such as SMAP~\cite{hassan_samp_2021} often use standard $A^*$ searching~\cite{hart1968formal} for this purpose.
The $A^*$ algorithm tends to generate deterministic shortest path for practice. However, humans usually move stochastically in the given scene.
To reflect the diversity of human path planning, we incorporate the standard $A^*$ algorithm with scene-aware random information concerning the diversity of human motion.

To begin with, we first discuss how to apply the standard $A^*$ algorithm into our scenario.
We first divide the whole 3D scene into the same set of non-overlapping discrete grids as in Section~\ref{subsec:interaction}.
We then define and calculate the cost function $f$ for each grid in the $A^*$ algorithm~\cite{hart1968formal} as:
\begin{align} \label{eq:astar}
    f(q) = g(q) + h(q);q \in \cN(p),
\end{align}
where $g(q)$ measures the cost for moving from the beginning point to grid $q$,
and $h(q)$ measures the cost between grid $q$ and the target grid, during searching points as the next step for $p$ in the neigbourhood $\cN(p)$.
To ensure obstacle-free paths, we further filter out inaccessible grids that might have collisions with the human body.
The collisions are detected via placing a cylinder model that approximates the volume of a human at each grid.
We show an example in the right of the Figure~\ref{fig:map} (a), where red stands for valid and blue stands for invalid.
After calculating $f$ for each grid and excluding invalid grids, an obstacle-free path connecting two human-scene interaction anchors can thus be obtained using the standard $A^*$ algorithm.
It is worth noting that the path obtained in this manner is deterministic and fixed for the same pair of two human-scene interaction anchors.

\begin{figure}[t]
	\centering
	\includegraphics[width=0.48\textwidth]{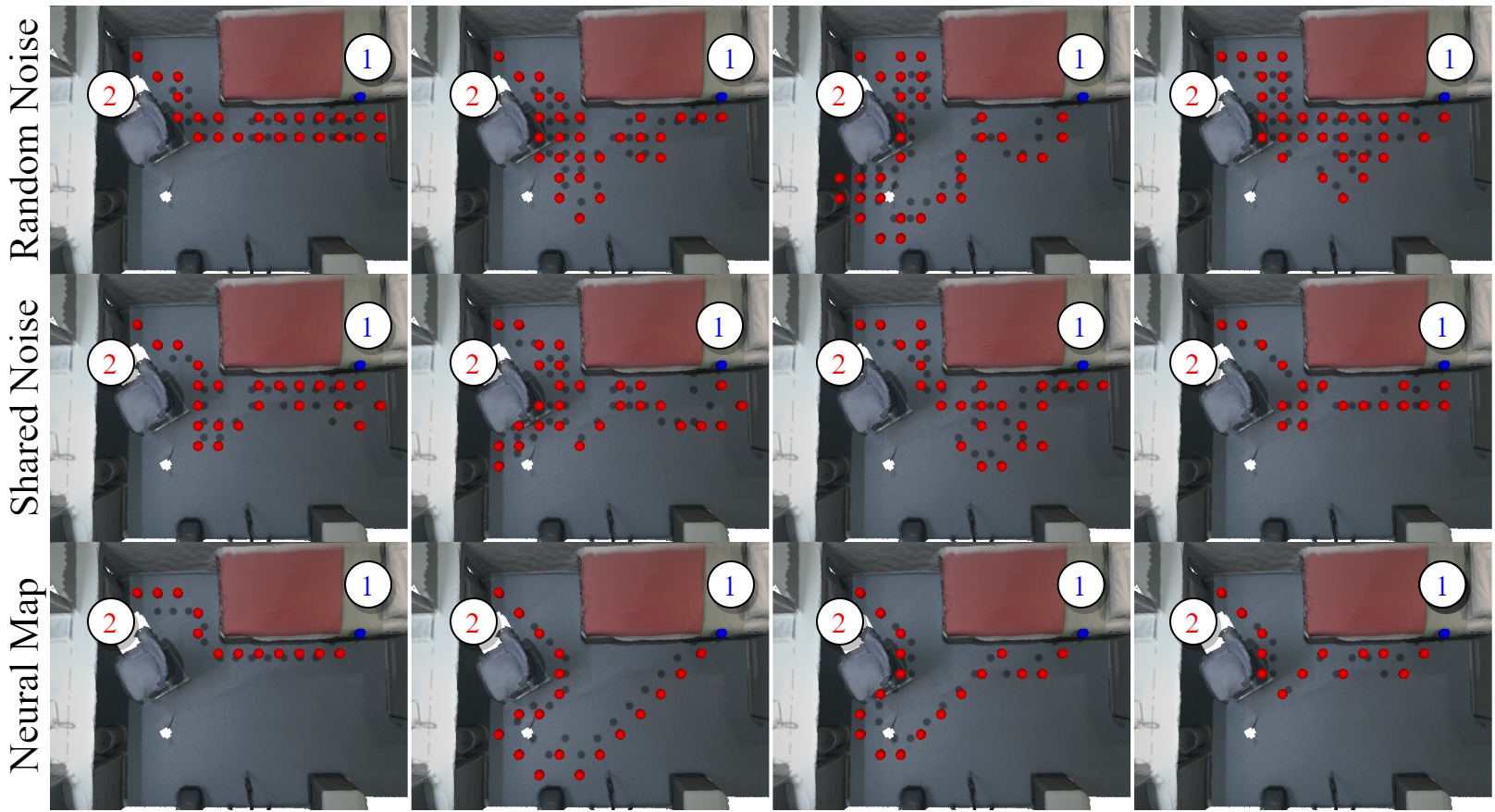}
	\small\caption{\textbf{Examples for diverse planning.} We sample different paths from $\textcircled{1}$ to $\textcircled{2}$ with different strategies. \textbf{Random Noise} means sampling different weights for each discrete gird. \textbf{Shared Noise} means all discrete girds share the same noise vector. \textbf{Neural Map} refers to probability generated by our Neural Mapper. The results demonstrate that our method is more effective in generating diverse and natural paths than simply adding randomly noise does.}
	\label{fig:planning_show}
	\vspace{-0.3cm}
\end{figure}

An intuitive solution to incorporate diversity in path planning is appending the cost function $f$ defined in Equation~\eqref{eq:astar} with a random noise term.
This strategy sounds feasible but fails to generate reasonable paths, which is demonstrated by the examples shown in the top two rows of Figure~\ref{fig:planning_show}.
To this end, we replace the random noise term with a controllable signal $m$ produced by another CVAE, referred as Neural Mapper.
For each grid $p$, Neural Mapper takes sampled latent code and the local scene context feature obtained via BPS~\cite{PLACE:3DV:2020,GRAB:2020} as the input and outputs the feasibility score for each neighbor grid $q \in \cN(p)$.
Based on the Neural Mapper, the cost function is updated as:
\begin{align} \label{eq:astar}
    f(p,q) &= g(q) + h(q) + (1 - m(p, q));q \in \cN(p).
\end{align}
The score of $m$ indicates the feasibility of moving from the current grid to this adjacent one so that we can build the cost as $1-m$ to reflect the moving guidance by our Neural Mapper.
The Neural Mapper is trained in a data-driven manner thus it can help the $A^*$ algorithm to generate diverse and reasonable paths.
We show several examples produced by Neural Mapper in the bottom row of Figure~\ref{fig:planning_show}.

Without complex manually designed conditions and constraints,
the proposed Neural Mapper equips the $A^*$ algorithm with the ability to find diverse obstacle-free paths in a flexible and generalizable way.
In Neural Mapper, we can easily change the characteristics of sampled paths by restricting the latent codes,
without hurting their naturalness and coherency.

%

\subsection{Motion Completion}~\label{sec:completion}
\begin{figure}[t]
	\centering
	\includegraphics[width=0.5\textwidth]{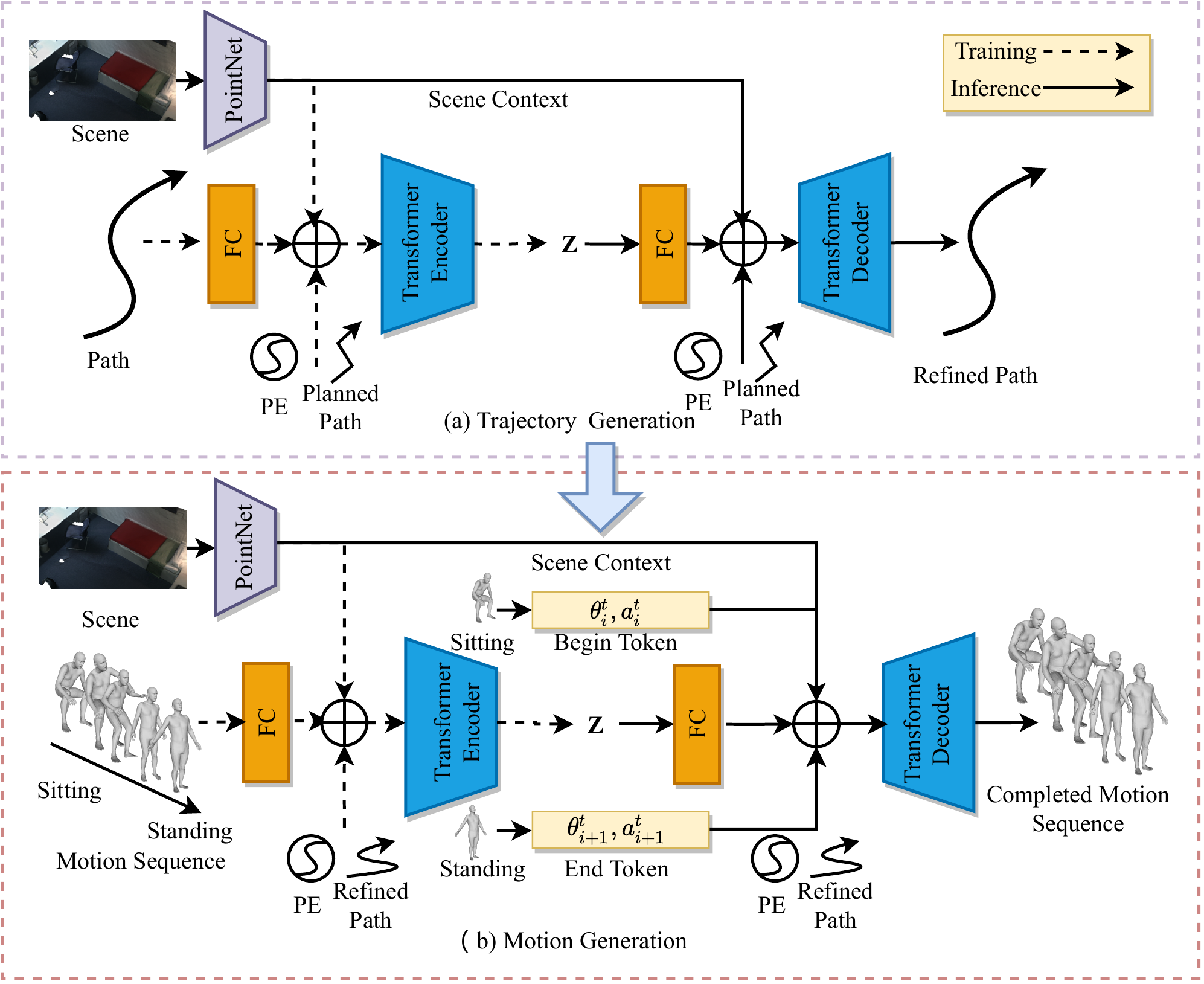}
	\small\caption{\textbf{Completion Network.} Motion completion network is composed of two modules, namely the Trajectory Refiner and the Motion Synthesizer. Trajectory Refiner reconstructs the motion trajectories under the guidance of planned path. The Motion Synthesizer takes the scene context, planned path, the pose and action label of human-scene interaction anchors as the inputs and generates body movements. The two-modules are trained altogether in an end-to-end manner.}
	\label{fig:motion}
	\vspace{-0.3cm}
\end{figure}

With the obstacle-free path obtained from path planning, we are now ready to complete the missing motions between consecutive human-scene interaction anchors.
As shown in Figure~\ref{fig:motion}, our motion completion network consists of two components, namely Path Refiner and Motion Synthesizer.
Although paths for human-scene interaction anchors are planned in Section~\ref{subsec:planning}, this Path Refiner accounts for the gap between diverse real human motions and the path formed by straight lines between the discrete grids.
Both the Path Refiner and the Motion Synthesizer follow the CVAE framework.
Specially, we apply Transformer~\cite{vaswani2017attention} as the basic architecture for both the encoder and decoder of these two networks to synthesize continuous and smooth motions.
Our motion completion network simultaneously synthesizes $M$ frame paths and body poses as~\cite{wang2021synthesizing, Wang_2021_CVPR}, instead of one-by-one in an auto-regressive manner~\cite{hassan_samp_2021, chuan2020action2motion, harvey2020robust}.

For Path Refiner, we take the scene context encoded by PointNet~\cite{qi2017pointnet} to synthesize the refined path.
The refined path is composed of pairs of the translation and orientation sequence$\{(t_1, \phi_1), ..., (t_M, \phi_M)\}$.
Following~\cite{petrovich21actor}, we introduce the positional encoding formed from sinusoidal functions which take time steps $t \in [1,...,M]$ as input to ensure the continuity and smoothness of the refined path.
Moreover, we leverage one more positional encoding obtained from the planned path by encoding each step of the planned path $(t_i, \phi_i)$ in Section~\ref{subsec:planning} by a fully connected layer, to ensure the refined path is still in the obstacle-free regions.
The effectiveness of this additional positional encoding is illustrated in Section~\ref{sec:exps},
where our Path Refiner further improves the diversity of synthesized motion.
The motion sequence with $M$ body poses $\{\theta_1, ...,\theta_M \}$ is completed by our Motion Synthesizer.
Same as the Path Refiner, we take the scene context encoded by the PointNet as the condition to complete these scene-aware motions.
The completed motions should fulfill two requirements, namely matching the paths produced by the Path Refiner and naturally transforming between the given human-scene interaction anchors.
To achieve this, the Motion Synthesizer at first takes the refined path as additional position encoding to guide motion synthesis, similar to the practice of Path Refiner.
For the motion transformation, we need to model the relationship between the given two human-scene interaction anchors and the potential motions that could be completed in our Motion Synthesizer.
Inspired by the practice of action token in~\cite{petrovich21actor}, which helps the transformer decoder to build up the relationship between synthesized motions and the given action, 
we encode the action labels and poses of human-scene interaction anchors by additional fully connected layers as learnable tokens and add them to the beginning and ending of the positional encoding respectively.
With these tokens, our Motion Synthesizer can directly build up this relationship between and synthesize reasonable and smooth motions.
Following these two steps, the motion completion network can generate natural motions for the given human-scene interaction anchors following the planned path.

\section{Experiments}
In this section, we first illustrate our experiment settings and metrics for evaluation. Then we discuss the effectiveness of the proposed framework. At last, we demonstrate the qualitative results in different scenes.

\subsection {Experimental Setting}

\noindent \paragraph{Implementation Details.}
All proposed CVAE models in the paper are optimized via ADAM~\cite{kingma2014adam} with learning rate set to $1\mathrm{e}{-4}$
All models are trained for $40$ epochs with batch size set to be $8$.
For better physical plausibility, we perform additional optimization used in~\cite{Hassan:CVPR:2021} and~\cite{wang2021synthesizing} to refine human-scene interaction anchors described in Section~\ref{subsec:interaction} and the completed motions described in Section~\ref{sec:completion}.
More details for the training scheduler and the optimization are included in the supplementary material.

\noindent\paragraph{Dataset.}
Following~\cite{PLACE:3DV:2020,PSI:2019,wang2021synthesizing}, we train our framework on PROX dataset~\cite{PROX:2019}.
We manually label the motions in PROX with action labels~(\ie~sit, lie, stand, walk, and squat) as the action condition. 
We do not conduct experiments on GTA-IM~\cite{cao2020long} and SAMP~\cite{hassan_samp_2021} as they do not provide reconstructed 3D real-world scenes. 
For the fair comparison, we follow the split of the train and test set as~\cite{PLACE:3DV:2020,PSI:2019,wang2021synthesizing} and synthesize human motions on the unseen scenes during training.
To demonstrate the generalization ability of the proposed framework, we further evaluate it on Matterport3D~\cite{Matterport3D}, which provides large-scale reconstructed 3D scenes.
Please be noted that our framework does not leverage Matterport3D for training.

\noindent \paragraph{Diversity Metric.}
We measure the diversity on synthesized human motions in three aspects, namely human-scene interaction anchors, planned paths, and completed motions.
To evaluate the diversity of the human-scene interacting anchors, we preform \textbf{K-Means} ($K=20$) clustering on the synthesized human-scene interaction anchors, following~\cite{Hassan:CVPR:2021}.
To be specific, we consider two types of the clusters. 
The first one considers all parameters $(\theta, t, \phi)$. The second one only considers translation $t$ and orientation $\phi$.
The diversity is measured as the entropy of the cluster sizes and the average distances between the clusters center and the samples belonging to it.
We evaluate path diversity by the standard deviation (\textbf{STD}) of distances between the paths from Neural Mapper and the ones from the standard $A^*$.
To fairly compare with previous works that manually assign anchors or target objects~\cite{PSI:2019, Hassan:CVPR:2021}, we evaluate the diversity of the synthesized human motions with the fixed human-scene interaction anchors.
To measure the ability of our motion completion network in generating diverse results, we do not introduce the diverse sampling strategies as~\cite{zhang2021we, yuan2020dlow}.
Following~\cite{hassan_samp_2021}, we calculate the Average Pairwise Distance (\textbf{APD}) on the SMPL-X parameters of synthesized motions to measure its diversity.

\noindent \paragraph{Naturalness Metric.}

We evaluate the naturalness of synthesized motions via \textbf{user study} and the physical plausibility. 
We ask users to compare our results against other methods and score them from 1 to 5 (the higher the better) as the results. 
Besides, we involve the \textbf{non-collision} score and \textbf{contact} score~\cite{PSI:2019, PLACE:3DV:2020,wang2021synthesizing} to measure the physical plausibility of synthesized motions between the 3D scenes.

\noindent \paragraph{Motion Metric.}
To evaluate the quality of the whole synthesized motions, we follow~\cite{hassan_samp_2021} to calculate the Frech\'{e}t Distance~(\textbf{FD}) between synthesized motions and ground-truth motions. 
This distance is computed using the parameters $P_i=(\theta_i, t_i, \phi_i)$ of each frame.

\subsection{Experimental Results}~\label{sec:exps}

\begin{table}
    \centering
    \setlength\tabcolsep{9pt}
    \caption{\textbf{Evaluation on human-scene interaction anchors.} We evaluate the diversity of the human-scene interaction anchors (Anchor, considering $\theta$, $t$, and $\phi$) and the placing (Position, considering only $t$ and $\phi$) with/without optimization post-process. $S$ means the sampling strategy based on pose relationship in Section~\ref{subsec:interaction}, and $R$ means our Placing Refiner.}\label{tab:anchors}
    \vspace{-0.2cm}
    \resizebox{0.48\textwidth}{!}{
    \begin{tabular}{l | c c| c c}
    \shline
    \multirow{2}{*}{Method}   & \multicolumn{2}{c|}{Anchor} &  \multicolumn{2}{c}{Position}\\
    & Entropy $\uparrow$ & Cluster $\uparrow$  & Entropy $\uparrow$ & Cluster $\uparrow$\\
    \hline
    Baseline~\cite{Hassan:CVPR:2021}          &  2.62 / 2.60  &  2.44 / 2.40   &  2.63 / 2.61  & 0.68 / 0.67  \\
    Baseline~\cite{Hassan:CVPR:2021} + S      &  2.74 / 2.73  &  2.55 / 2.53   &  2.69 / 2.68  & 0.79 / 0.78 \\   
    Baseline~\cite{Hassan:CVPR:2021}+ S + R  &  \textbf{2.77} / \textbf{2.73}  &  \textbf{2.57} / \textbf{2.53}   &  \textbf{2.72} / \textbf{2.70}  & \textbf{0.83} / \textbf{0.80}   \\       
    \shline
    \end{tabular}}
\end{table}

\begin{figure}[t]
	\centering
	\includegraphics[width=0.5\textwidth]{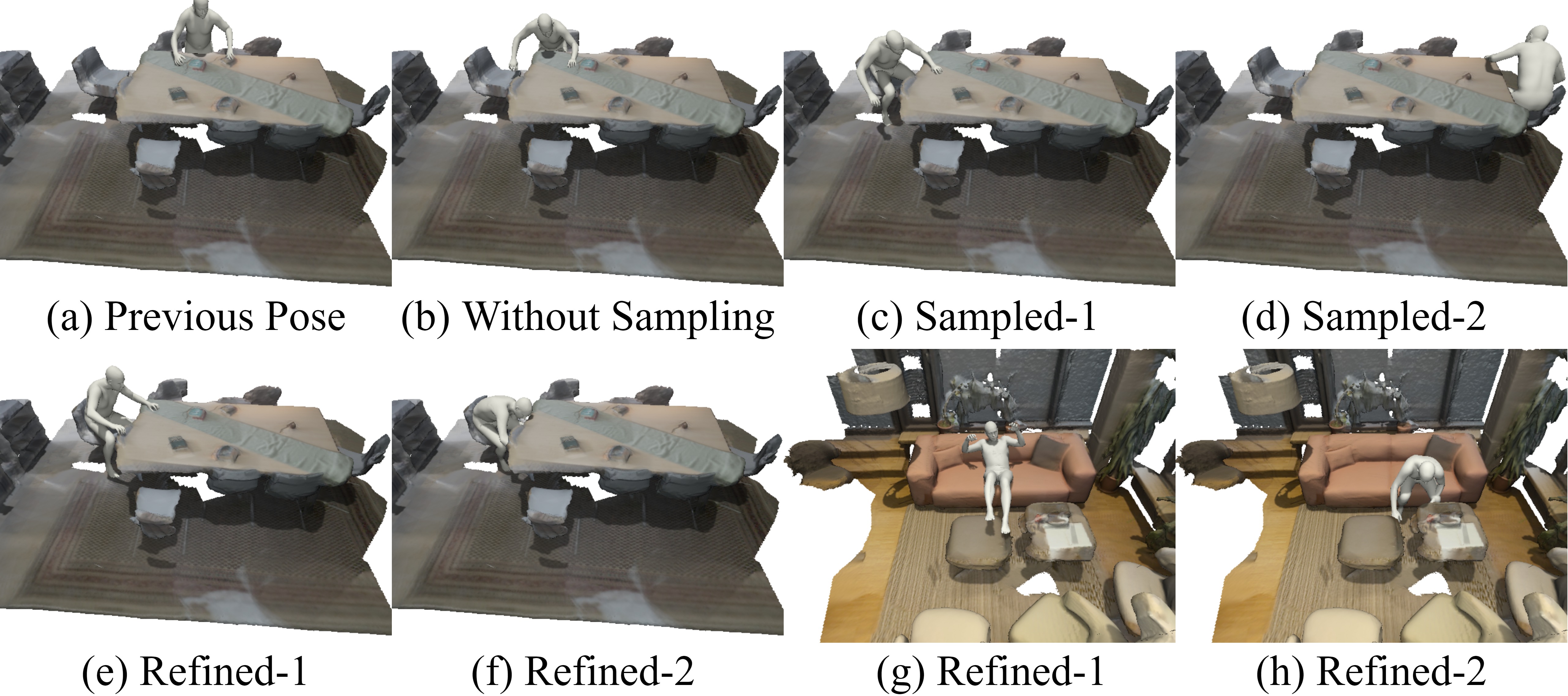}
	\vspace{-0.5cm}
	\small\caption{\textbf{Placing Results.} The first row shows the results with and without the pose related sampling strategy. The second row shows the results where our Place Refiner and the optimization post-processing in~\cite{Hassan:CVPR:2021} works together.}
	\label{fig:placing_show}
	\vspace{-0.3cm}
\end{figure}

\begin{figure*}[t]
        \centering
        \includegraphics[width=0.96\textwidth]{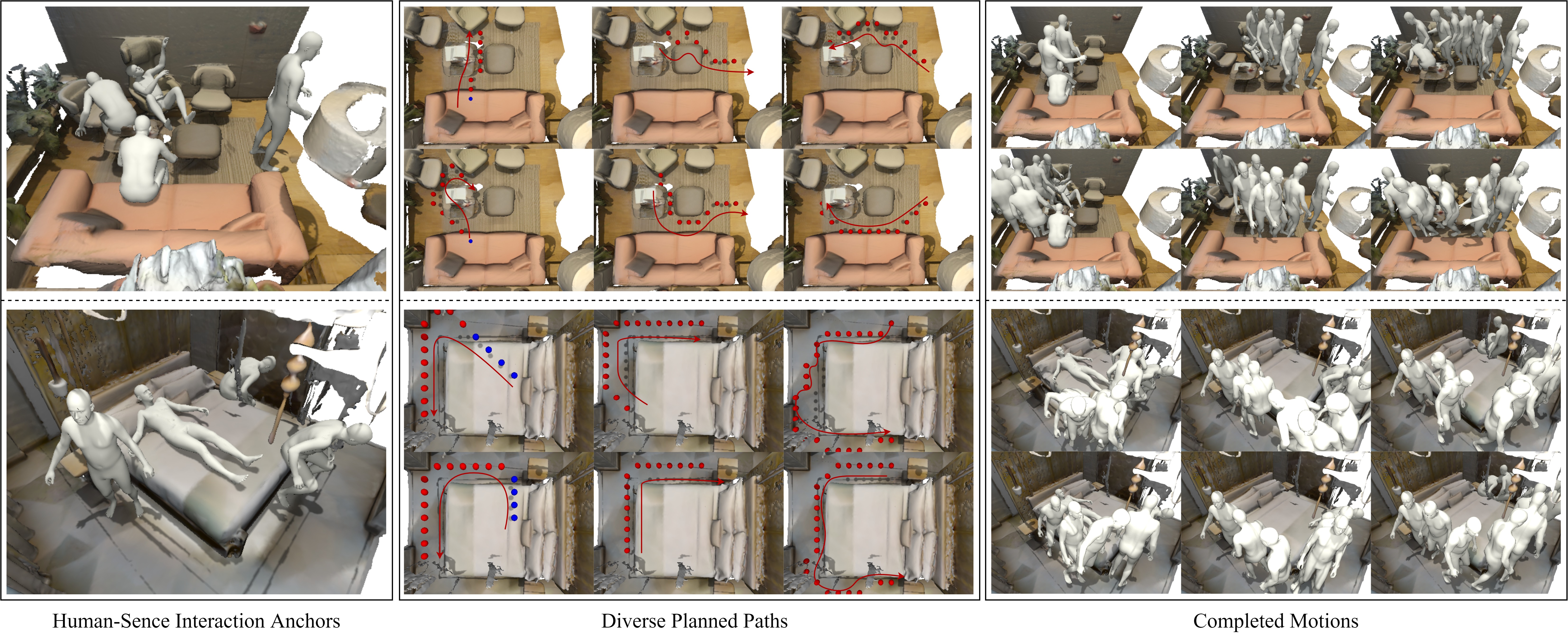}
        \small\caption{\textbf{Qualitative Results.} In this figure, we show the output of each stage in our framework. The first row is synthesized on PROX dataset, and the second one is on Matterport3D. Our framework can synthesize motions with diverse interactions to the given scene.}
        \label{fig:results}
\end{figure*}

In this section, we show quantitative results on PROX dataset. The quantitative results on Matterport3D dataset are included in our supplementary materials. We also show qualitative results on these two datasets in this section.

\noindent\paragraph{Human-Scene Interaction Anchors.}
We first show the diversity of synthesized human-scene interaction anchors in Table~\ref{tab:anchors}. 
For this evaluation, we sample 100 poses for each action and employ the placing strategy in POSA~\cite{Hassan:CVPR:2021} as our baseline.
As shown in the table, the position related sampling process and the Pose Refiner can both improve the diversity of interaction anchors. 
We further show the effectiveness of these two process in Figure~\ref{fig:placing_show}. 
Examples shown in this figure are all generated from the action label ``sit''.
(a) shows the first placed poses.
Without the position related sampling, (a) and (b), which have the same action label, are placed close to each other. 
(c) and (d) are the generated anchors using our position related sampling. 
It is revealed that they interact with different objects in the scene.
The second row demonstrates the result pairs ((d) V.S. (e) and (f) V.S. (g)), which are produced by our Place Refiner works and optimization post-process~\cite{Hassan:CVPR:2021}.
Using the diverse translations and orientations as initialization states, the optimization algorithm can produce diverse optimal solutions.
In the supplemental material, we show comparison with previous works~\cite{PSI:2019,PLACE:3DV:2020} that are extended to synthesize specific actions using our action condition.

\begin{table}
    \centering
    \setlength\tabcolsep{9pt}
    \caption{\textbf{Evaluation on diverse planning module}. We compare against the standard $A^*$ algorithm and methods with sampled random noises. The metrics show the diversity and naturalness of the planned paths. }\label{tab:planning}
    \vspace{-0.3cm}
    \resizebox{0.48\textwidth}{!}{
    \small\begin{tabular}{l | ccccc| c}
    \shline
    Method              &  1/6 $\uparrow$  &  1/3 $\uparrow$ & 1/2 $\uparrow$  & 2/3 $\uparrow$ & 5/6 $\uparrow$ & User Study $\uparrow$ \\
    \hline
    Standard            &  0  &   0  &  0   &  0 & 0  &  4.31(0.48)\\
    Random Noise        & 0.346  &  0.566    &  0.628  &  0.523  &  0.324 & 2.52(0.53) \\
    Shared Noise        & 0.297  &  0.483    &  0.603  &  0.485  &  0.281 &   3.52(0.45) \\
    \hline
    Ours                & 0.286  &  0.446    &  0.508  &  0.415  &  0.233  & 4.27(0.52)\\       
    \shline
    \end{tabular}}
\end{table}

\noindent\paragraph{Planning.}
We compare the diversity and naturalness of the planned path. 
For the evaluation of diversity, We compute the standard deviation of the distances between sampled paths and the paths produced by standard $A^*$ . 
In practice, we calculate distances between the discrete points on the paths, which are set as 1/6, 1/3, 1/2, 2/3, and 5/6 of the sample paths.
We also show the results of the user study to reflect the naturalness of the planned paths. The number of samples is set to be $50$ for each method.
The evaluation results are shown in Table~\ref{tab:planning}. 
Standard $A^*$ algorithm, which is used in~\cite{hassan_samp_2021} only produces the deterministic path for practice while the proposed Neural Mapper can generate diverse paths with similar naturalness as the standard A$*$ does.
On the other hand, two methods using random noises cannot produce natural results, although they generate more diverse paths than ours.
Similar results are also demonstrated in Figure~\ref{fig:planning_show}.
Compared with the methods using random noises, Neural Mapper can provide consistent and reasonable guidance for the similar local scene context to avoid unnatural moving. 
Moreover, Neural Mapper can cope with other manual constraints such as avoiding passing a certain region. We will discuss it in our supplementary materials.

\noindent\paragraph{Motion Synthesis.}

\begin{table}
	\centering
	\setlength\tabcolsep{9pt}
	\caption{\textbf{Evaluation on motion completion module.} We mainly evaluate the models in two aspects. FD is used to show the completion ability. APD is used to evaluate motion diversity. We compare our framework with several state-of-the-art methods. Specially, ``w/ OPT'' and ``w/o OPT'' refer to the results obtained with/without optimization post-process~\cite{wang2021synthesizing}. ``Ours*'' means our motion completion network without the Path Refiner.}\label{tab:completion-1}
	\vspace{-0.3cm}
	\resizebox{0.48\textwidth}{!}{\small\begin{tabular}{l |  c c | c c  }
			\shline
			\multirow{2}{*}{Method} &  \multicolumn{2}{c|}{FD $\downarrow$} & \multicolumn{2}{c}{APD $\uparrow$} \\
			& w/o OPT & w/ OPT  & w/o OPT & w/ OPT\\
			\hline
			SA-CSGN~\cite{Wang_2021_CVPR}     & 176.20 & 175.28  & 2.13 &  2.15   \\

			Wang et.al.~\cite{wang2021synthesizing}   & 121.22 & 120.01    & 0.00  & 0.00  \\
			SAMP~\cite{hassan_samp_2021}           & 115.34 & 114.22 &  2.56   &  2.57   \\
			\hline
			Ours*           & 126.46 & 124.52  &   2.46  & 2.46  \\    
			Ours                 & \textbf{112.74}  & \textbf{111.65} & \textbf{2.77}    & \textbf{2.78} \\       
			\shline
	\end{tabular}}
    \vspace{-0.5cm}
\end{table}

In this subsection, we compare with other advances on scene-aware motion synthesis~\cite{wang2021synthesizing, hassan_samp_2021, Wang_2021_CVPR}.  
We use the official model of~\cite{wang2021synthesizing} trained on PROX dataset and extend~\cite{hassan_samp_2021} and~\cite{Wang_2021_CVPR} to PROX for fair comparison.
In Table~\ref{tab:completion-1}, we first compare against these methods using \textbf{FD} and \textbf{APD} for the motion quality and diversity. 
Firstly, for the comparison of \textbf{FD}, we sample 500 motion sequences which begin with the same action, as well as 500 motions which finish the same action.
Besides, for the comparison of \textbf{APD}, we sample $100$ pairs of human-scene interaction anchors and synthesize $10$ motions for each pair. 
It is revealed that our method achieves the best results against other methods. 
All the comparison results show that our method can synthesize more diverse and natural motions than other methods do.
In the supplementary material, we first compare more naturalness results between these methods, \eg physical compatibility and user study. Then we further discuss our design choice on the motion completion network, including the effectiveness of the Path Refiner and the positional encoding based on planned paths.

%

\noindent\paragraph{Qualitative Results.}

We show more qualitative results of the proposed method on the PROX~\cite{PROX:2019} and Matterport3D~\cite{Matterport3D} in Figure~\ref{fig:results}. 
We show all three aspects of the synthesized scene-aware motions, including the human-scene interaction anchors, diverse planned paths, and completed motions. 
These results demonstrate that our framework can synthesize diverse human motions in the specific scene contexts for the given target action sequence. 
More qualitative results are included in the supplemental material.
\section{Conclusion}
In this paper, we focus on synthesizing diverse and natural human motions in the given scene environment guided by target action sequence.
We decompose the diversity of scene-aware human motions into three levels, namely the diversity of action-conditioned human-scene interactions, the diversity of obstacle-free paths, and the diversity of body movements.
To comprehensively leverage the inherent diversity of human motions, we propose a novel hierarchy framework with each component accounting for each level of the diversity.
Thanks to the effective decomposition of diversity and elaborated designed modules, our framework is able to produce various vivid human motions in the scene across all three levels with improved efficiency and generality.
Furthermore, the factorized design of our framework make it can be easily incorporated into other human motion synthesizing frameworks.

\section{Supplementary Materials}
\subsection{Implementation Details}
In this section, we first discuss the detailed structures of the CVAE models used in our framework. Then we discuss the training and inference details of these models.

\subsubsection{Network Architecture.}
The encoders and decoders of the action conditioned pose generator in Section 3.2, Place Refiner in Section 3.2, and Neural Mapper in Section 3.3 share the exactly same architectures, which are all two-layer Multilayer Perceptron (MLP).
The encoder takes the $256$-dim features encoded by the fully-connected layers and predicts the mean $\mu \in \mathbb{R}^{32}$ and the standard deviation $\sigma \in \mathbb{R}^{32}$ for a Gaussian Distribution.
We sample the latent code $z$ from this distribution for the decoder during training.

For the motion completion network in Section 3.4, we use the Transformer~\cite{vaswani2017attention} as the basic structure as~\cite{petrovich21actor}. 
To be specific, we use two fully connected layers to encode all inputs to $256$-dimension features. 
The encoder predicts the mean $\mu \in \mathbb{R}^{32}$ and the variance $\sigma \in \mathbb{R}^{32}$ for a Gaussian Distribution as the CVAE model for action conditioned poses.
Following~\cite{petrovich21actor}, we set $8$ layers of the Transformer network for the encoder and decoder.

\subsubsection{Training and Inference Details.}
\noindent\paragraph{Scene-Agnostic Pose Synthesis.}
Firstly, we show how to train the CVAE model for scene-agnostic pose synthesis in Section 3.2.
As the standard VAE~\cite{kingma2013auto} model, the training objective consists of two parts. 
The first one is the reconstruction loss between the reconstructed human poses and the input human pose. 
The other objective is Kullback-Leibler~(KL) Divergence between the Gaussian Distribution $Q(z|\mu, \sigma)$, where $\mu$ and $\sigma$ are predicted by the encoder, and the standard Gaussian Distribution $N(0, I^2)$. 

\noindent\paragraph{Place Refiner.}
The Place Refiner takes the placed body poses, and the scene contexts encoded by the PointNet~\cite{qi2017pointnet} as inputs and predict the offset $\Delta t_i, \Delta o_i$ for the sampled $\bar t_i, \bar o_i$. 
To train this network, we first build up the discrete candidates following the same procedure of scene-conditioned anchor placing in Section 3.2 for each scene in our training set. 
For practice, we split each scene into non-overlapping discrete grids uniformly as translation candidates and then uniformly sample eight different orientations paralleling with the ground plane as the orientation candidate.
Each pose in the given scene is neighbor to four-position candidates.
We assign each pose in the training set to one randomly sampled neighbor position candidate and one orientation candidate of this position candidate.
Then we predict the offset from these candidates to the original translation and orientation for this pose.
Similar as~\cite{kingma2013auto}, the training objective is the reconstruction loss and the KL-Divergence.

\noindent\paragraph{Neural Mapper.}
The input of Neural Mapper includes the local context encoded by BPS~\cite{PLACE:3DV:2020} and the human moving direction in this local context, which is obtained from ground-truth moving paths of PROX~\cite{PROX:2019} dataset.
To train this model, we first split the motions in the training set of the PROX dataset into different 60 frame sequences. 
The local context is cropped as a $2m \times 2m \times 2m$ cubic cage at the motion center as~\cite{PLACE:3DV:2020}. 
To compute BPS features, we uniformly sample a set of $N_b=10^4$ basis points within the unit sphere at the center of the local context and then normalize the local scene context into the same unit sphere. 
The final BPS feature is the concatenation of these minimal distances $\vx_s \in \mathbb{R}^{N_b \times 1}$ between the sampled unit sphere and normalized scene context.
In practice, we use two additional fully-connected layers to further encode $\vx_s$ as the local context feature. 
Similar to the standard VAE~\cite{kingma2013auto}, the training objective of Neural Mapper consists of a KL-Divergence term and a reconstruction term. 
Specially, we norm the moving direction between the beginning and ending points of the motion sequence to $[0, 1]$ as the reconstruction target during training.
The reconstruction term estimates the residuals between this normalized moving direction and the expectation of the estimated direction distribution.

\noindent\paragraph{Path Refiner and Motion Synthesizer.}
We train our Path Refiner and Motion Synthesizer together in an end-to-end manner. 
Both two models synthesize $M=60$ frames of paths or motions. 
Similarly, the training objective consists of the reconstruction loss on the synthesized paths or motions, as well as the KL-Divergence. 
Specially, we do not use the planned path obtained from Section 3.3 in training the Path Refiner.
Instead, we use the directions pointing from the beginning to the ending point of the motion as the planned path for practice.
The reason mainly lies in two aspects. 
The first one is that repeat running of path planning module to obtain planned paths is not efficient in training.
The second one is that the shortest path from the planning module is similar to the straight line in short-term motions.

During the inference stage, we find that directly synthesizing motions from two consecutive anchors lead to unstable results. 
We believe it is majorly caused by the variance of the lengths of the planned paths. 
To resolve this, we first split the planned paths into several pieces with equal length. 
Each split point is then assigned with an intermediate status action label.
Using the new action labels, the intermediate anchors can be produced following the same method as placing human-scene interaction anchors in Section 3.2. 
Given the new anchors and the split paths, our motion completion network synthesizes human motions for each piece and then connects them together as the integrated motion. 
For practice, we insert the motions with random poses conditioned on ``walking'', ``standing'', and `` squatting'' action.

\subsubsection{Optimization.}
We perform optimizations to improve the motion quality with the motion and physics constraints. For example, the human should walk on the floor with smooth motions.
We conduct the optimization in~\cite{Hassan:CVPR:2021} and~\cite{wang2021synthesizing} for human-scene interaction anchors and motion sequences, respectively. 
For better human-scene interaction anchors, we use the objective functions defined in~\cite{Hassan:CVPR:2021}, that consist of the affordance loss for contacting the specific body parts to the given scene~(\eg~foot to the floor), 
penetration loss for the reasonable physical relationship between body meshes and the reconstructed SDF (sign distance field), 
and the regularization to keep the optimized pose close to the initial pose. 
We optimize all these human-scene interaction anchors for $10$ iterations with $1\mathrm{e}{-3}$ learning rate, using L-BFGS~\cite{LiuN89} algorithm as~\cite{Hassan:CVPR:2021}. 
For the synthesized motion obtained in Section 3.4, we follow their optimization objective functions~\cite{wang2021synthesizing} for foot location, environment, and motion smoothness to improve the motion quality.
We optimize all our motions for $100$ iterations with $1\mathrm{e}{-2}$ learning rate, using ADAM~\cite{kingma2014adam} algorithm as~\cite{wang2021synthesizing}.

\subsection{Experiments}

\begin{table}
    \centering
    \setlength\tabcolsep{9pt}
    \caption{\textbf{Evaluation on naturalness of synthesized motions on PROX~\cite{PROX:2019}.} We measured this by physical plausibity (non-collision and contact score) as well as user study.Specially, w/ and w/o opt means the results with/without optimization post-process~\cite{wang2021synthesizing}. ``Ours*'' means our motion completion network without the Path Refiner.}\label{tab:completion-2}
    \resizebox{0.48\textwidth}{!}{\small\begin{tabular}{l | c c | c c | c }
    \shline
    \multirow{2}{*}{Method}   & \multicolumn{2}{c|}{Non-Collision $\uparrow$  } &  \multicolumn{2}{c|}{Contact  $\uparrow$}  & User Study $\uparrow$\\
    & w/o opt & w/ opt & w/o opt& w/ opt \\
    \hline
    SA-CSGN~\cite{Wang_2021_CVPR}   &  92.37     &   98.21   &  97.16 &  98.72 & 2.74(0.97)\\
    Wang et.al.~\cite{wang2021synthesizing}    &  93.88     &   98.72 &     98.31 &  99.35     & 3.42(1.06)\\
    SAMP~\cite{hassan_samp_2021}        &  94.92     &  99.31    &    98.18 &  99.32    & 3.46(0.96) \\
    \hline
    Ours*               &  94.52     &   99.28  &   98.22  &  99.27  & 3.28(0.94)\\ 
    Ours                &  \textbf{95.93}     &   \textbf{99.61}  &     \textbf{98.33} &  \textbf{99.35}  & \textbf{3.68(0.84)}\\       
    \shline
    \end{tabular}}
\end{table}

\begin{table}
    \centering
    \setlength\tabcolsep{9pt}
    \caption{\textbf{Evaluation on human-scene interaction anchors for Matterport3D~\cite{Matterport3D}.} We evaluate the diversity of the human-scene interaction anchors (Anchor, considering $\theta$, $t$, and $\phi$) and the placing (Position, considering only $t$ and $\phi$) with/without optimization post-process. $S$ means the sampling strategy based on pose relationship in Section 3.2, and $R$ means our Placing Refiner.}\label{tab:anchors}
    \vspace{-0.2cm}
    \resizebox{0.48\textwidth}{!}{
    \begin{tabular}{l | c c| c c}
    \shline
    \multirow{2}{*}{Method}   & \multicolumn{2}{c|}{Anchor} &  \multicolumn{2}{c}{Position}\\
    & Entropy $\uparrow$ & Cluster $\uparrow$  & Entropy $\uparrow$ & Cluster $\uparrow$\\
    \hline
    Baseline~\cite{Hassan:CVPR:2021}          &  2.54 / 2.50 &  2.45 / 2.44   &  2.51 / 2.50  & 0.58 / 0.56  \\
    Baseline~\cite{Hassan:CVPR:2021}   + S      &  2.63 / 2.61  &  2.53 / 2.53   &  2.54 / 2.54  & 0.64 / 0.65 \\   
    Baseline~\cite{Hassan:CVPR:2021}   + S + R  &  \textbf{2.70} / \textbf{2.68}  &  \textbf{2.59} / \textbf{2.58}   &  \textbf{2.68} / \textbf{2.66}  & \textbf{0.72} / \textbf{0.72}   \\       
    \shline
    \end{tabular}}
\end{table}

\begin{table}
	\centering
	\setlength\tabcolsep{9pt}
	\caption{\textbf{Evaluation on synthesized motion for Matterport3D~\cite{Matterport3D}.} Comparison on \textbf{APD}, \textbf{non-collision} score and \textbf{contact} score on Matterport3D dataset. Specially, ``w/ OPT'' and ``w/o OPT'' refer to the results obtained with/without optimization post-process~\cite{wang2021synthesizing}. ``Ours*'' means our motion completion network without the Path Refiner.}\label{tab:completion-2}
	\vspace{-0.3cm}
	\resizebox{0.48\textwidth}{!}{\small\begin{tabular}{l | cc | cc | cc }
			\shline
			\multirow{2}{*}{Method}  & \multicolumn{2}{c|}{APD $\uparrow$}  & \multicolumn{2}{c|}{Non-Collision $\uparrow$  } &  \multicolumn{2}{c}{Contact  $\uparrow$}  \\
			& w/o OPT & w/ OPT  & w/o OPT & w/ OPT & w/o OPT & w/ OPT \\
			\hline
			SA-CSGN~\cite{Wang_2021_CVPR}  & 2.24 &  2.26 & 91.51 & 99.08 & 99.11  & 99.33   \\

			Wang et.al.~\cite{wang2021synthesizing}   & 0.00  & 0.00  & 93.78 &  99.42 & 99.32  & 99.35 \\
			SAMP~\cite{hassan_samp_2021}    & 2.46   &  2.48  & 94.35 & 99.32 & 99.28  & 99.35\\
			\hline
			Ours*        &   2.34  & 2.38  & 94.12 & 99.08  & 99.18 &  99.32 \\    
			Ours         & \textbf{2.57}    & \textbf{2.60} & \textbf{95.72} & \textbf{99.42} &  \textbf{99.32} & \textbf{99.36}\\       
			\shline
	\end{tabular}}
\end{table}

\noindent\paragraph{Naturalness Results on PROX.}
Then we compare the naturalness of these methods in Table~\ref{tab:completion-2}. 
For the physical plausibility, we use the same motion as the comparison in Table.3 of our paper.
For user study, we randomly sample motions with $2$, $4$, and $8$ different target actions.
All the comparison results show that our method can synthesize more natural motions than other methods do. 
Especially, our method achieves better results without the optimization post-process, because of the guidance from planned obstacle-free paths. 
The comparison between the last two rows shows that the guidance of the proposed Path Refiner is advantageous in synthesizing natural motions.

\noindent\paragraph{Results on Matterport3D.}
We show the quantitative results on Matterport3D dataset~\cite{Matterport3D}.
The sampling strategies are the same as our experiments on PROX dataset.
We first perform \textbf{K-Means} ($K=20$) and evaluate the obtained human-scene interaction anchors on Matterport3D with the entropy of cluster sizes and the average distance between the cluster center and the samples belong to it.

As shown in Table~\ref{tab:anchors}, our method enhance the diversity of the anchors for motion synthesis. 
Besides, we evaluate the synthesized motion on Matterport3D via the \textbf{APD}, \textbf{Non-Collision} score and \textbf{Contact} score.
The results are listed in Table~\ref{tab:completion-2}.
It is revealed that our method can synthesize better results than previous methods with better diversity and physical plausibility.
Besides, the Path Refiner still can improve the diversity and naturalness on this dataset.

\subsection{Further Discussion}
\begin{figure}[t]
	\centering
	\includegraphics[width=0.5\textwidth]{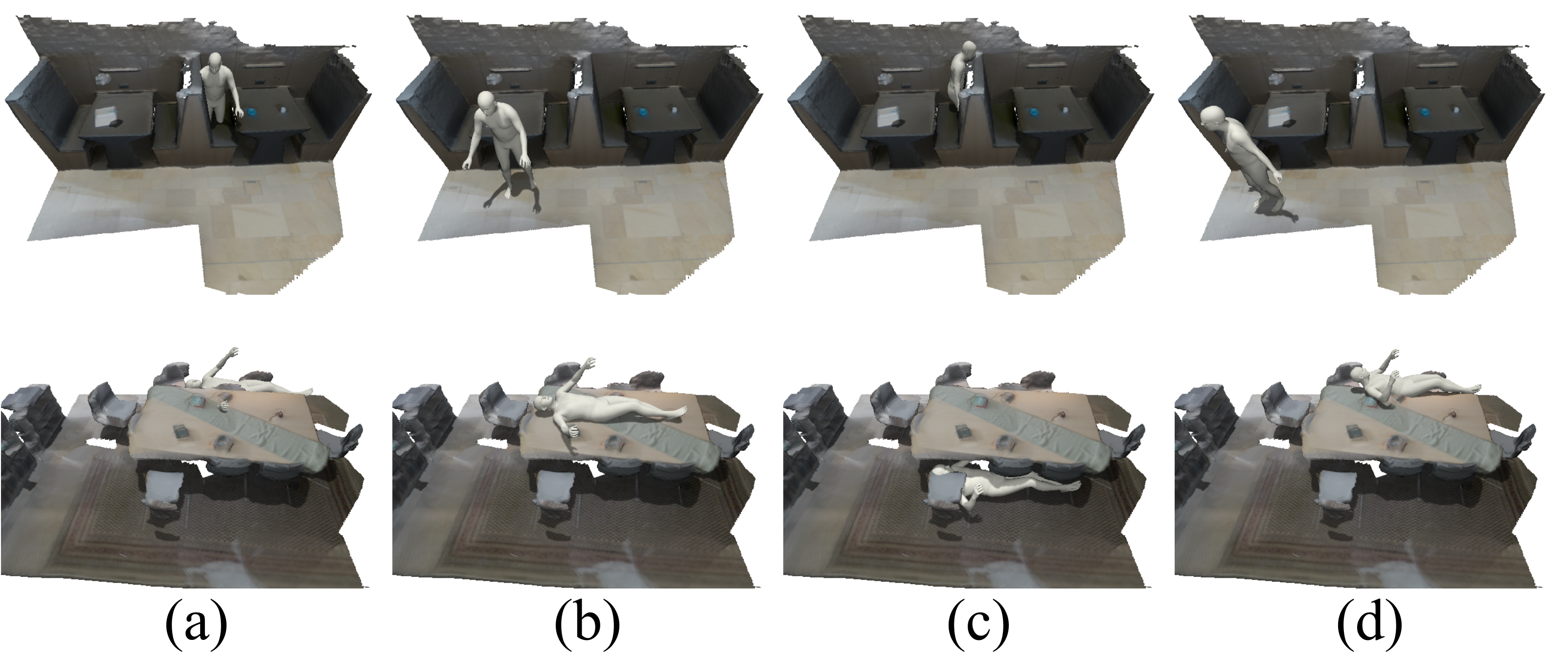}
	\small\caption{\textbf{Failure cases of the scene-centric paradigm.}
	We use the action label as additional condition to extend previous scene-centric paradigms~\cite{PSI:2019,PLACE:3DV:2020}. The first and third column shows results obtained from \cite{PSI:2019} and \cite{PLACE:3DV:2020}, respectively. The second and forth columns shows the results generated by our framework with the same pose as the first and third columns, respectively.
	}
	\label{fig:failure_case}
	\vspace{-0.3cm}
\end{figure}

In this section, we first discuss the reason for using the human-centric paradigm, for human-scene interaction anchors. 
The human-centric paradigm means we place the sampled poses to the positions which match the physical structure of these poses.
Then we show how to use our Neural Mapper to work with other manually set constraints. 
At last, we show the influence of the planned path on motion synthesis.

\noindent\paragraph{Human-Scene Interaction Anchor.} 
Previous works~\cite{PSI:2019, PLACE:3DV:2020} of synthesizing human-scene interaction anchors aim to explore the influence of scene context to place human pose in the given scene and neglect the action labels. 
Intuitively, we can incorporate these action labels as an additional condition and incorporate them into their frameworks to synthesize poses.
However, as shown in Figure~\ref{fig:failure_case}, simply extending the previous works cannot guarantee to synthesize the physically plausible poses with the given actions and scene contexts.
We believe it is due to the reason that these methods do not build up the relationship between the action and the scene context explicitly.
For example, method~\cite{PSI:2019} directly uses the pooled 2D image features as the condition and ignores the relationship between the spatial information and the action.
Another method~\cite{PLACE:3DV:2020} first samples different positions to build up the BPS and then synthesize different poses. 
However, the poses for each action have their specific physical structure and match different scene structures. 
It is difficult to find suitable places for the poses conditioned on the given action label, as shown in the Figure~\ref{fig:failure_case}. 
Instead, the human-centric paradigm proposed by us can effectively leverage the explicit relationship between the synthesized 3D human and the scene structure (\eg physical and semantic structures) and thus makes the whole placing process more controllable.

\begin{figure}[t]
	\centering
	\includegraphics[width=0.5\textwidth]{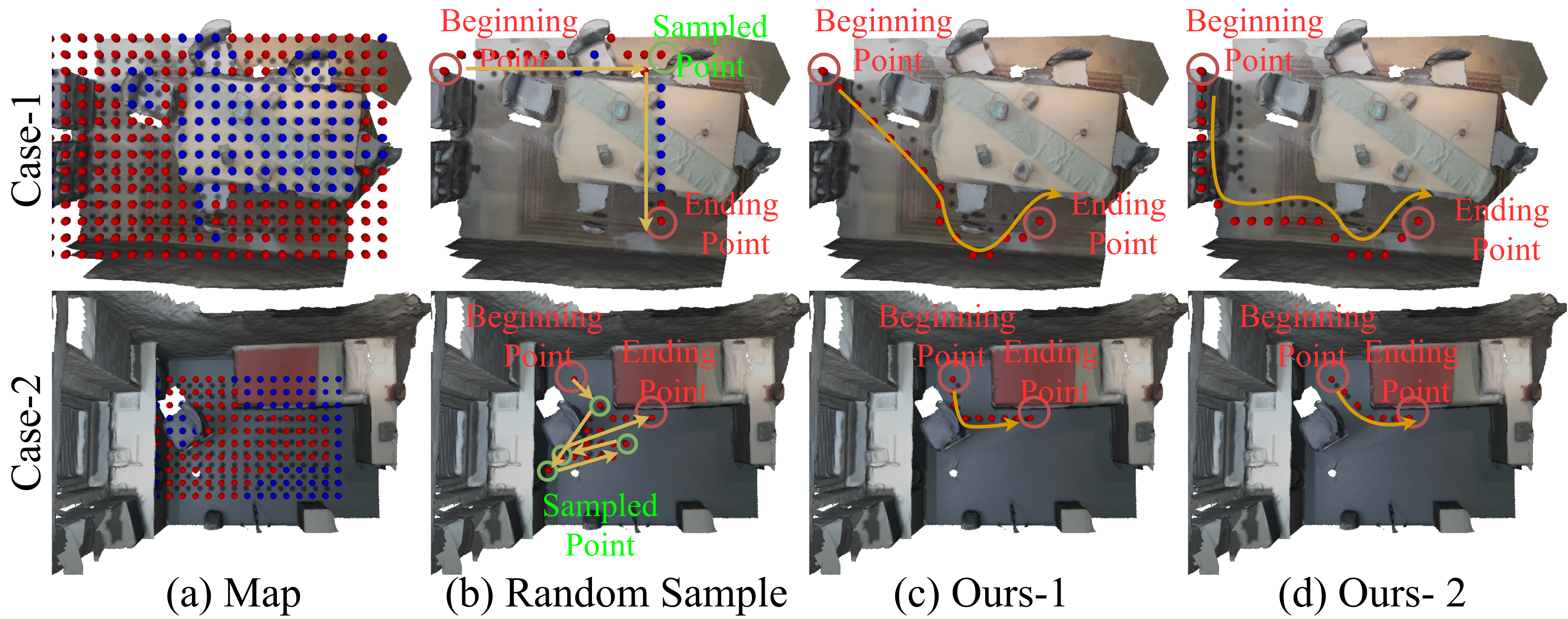}
	\small\caption{\textbf{ Comparison with randomly sampled intermediate points.} Our method can plan diverse and natural paths without complex manual constraints.}
	\label{fig:random_sampling}
	\vspace{-0.3cm}
\end{figure}

\begin{figure}[t]
	\centering
	\includegraphics[width=0.5\textwidth]{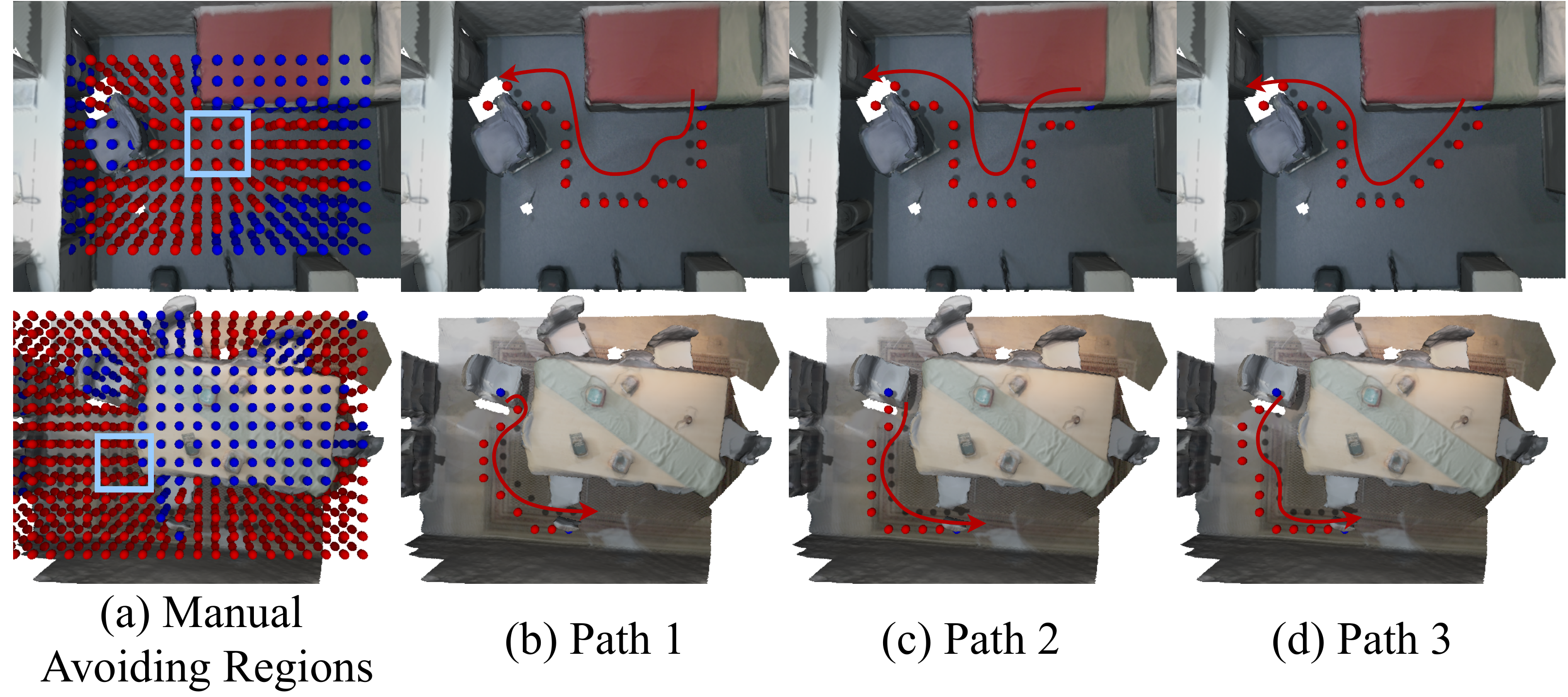}
	\small\caption{\textbf{Neural Map with manual constraints.} We change the valid grids in the blue box as the manual avoiding regions. These regions are the shortest path from these two points. We find that our Neural Map can work with this constraint to sample different planned paths.}
	\label{fig:manual_avoid}
	\vspace{-0.3cm}
\end{figure}

\noindent\paragraph{Neural Mapper}

Several failure cases generated from randomly sampling intermediate points are included in Figure~\ref{fig:random_sampling}.
In the first row, when sampled intermediate points and the ending points are obstructed, the original $A^\star$ algorithm can not find paths for these points. We adjust $A^\star$ by allowing to search paths in the obstacle regions , and $A^\star$ only produces impractical paths crossing the table as the first row of Figure~\ref{fig:random_sampling}.
In the second row, random sampled points can also lead to unnatural zigzag paths.
One may argue that these failures can be avoided via complex constraints used in previous methods~\cite{vsvestka1998probabilistic,carpin2006randomized}.
However, the proposed Neural Mapper provides an \textbf{automatic and data-driven} way to embed semantic information into natural and diverse path planning, without complex constraints.
Besides, our Neural Mapper also can work with manually set constraints, such as avoiding passing a certain region. 
We show the planned results in Figure~\ref{fig:manual_avoid}. 
It is revealed that our method can still produce natural and diverse paths under such constraints.

\begin{figure}[t]
	\centering
	\includegraphics[width=0.48\textwidth]{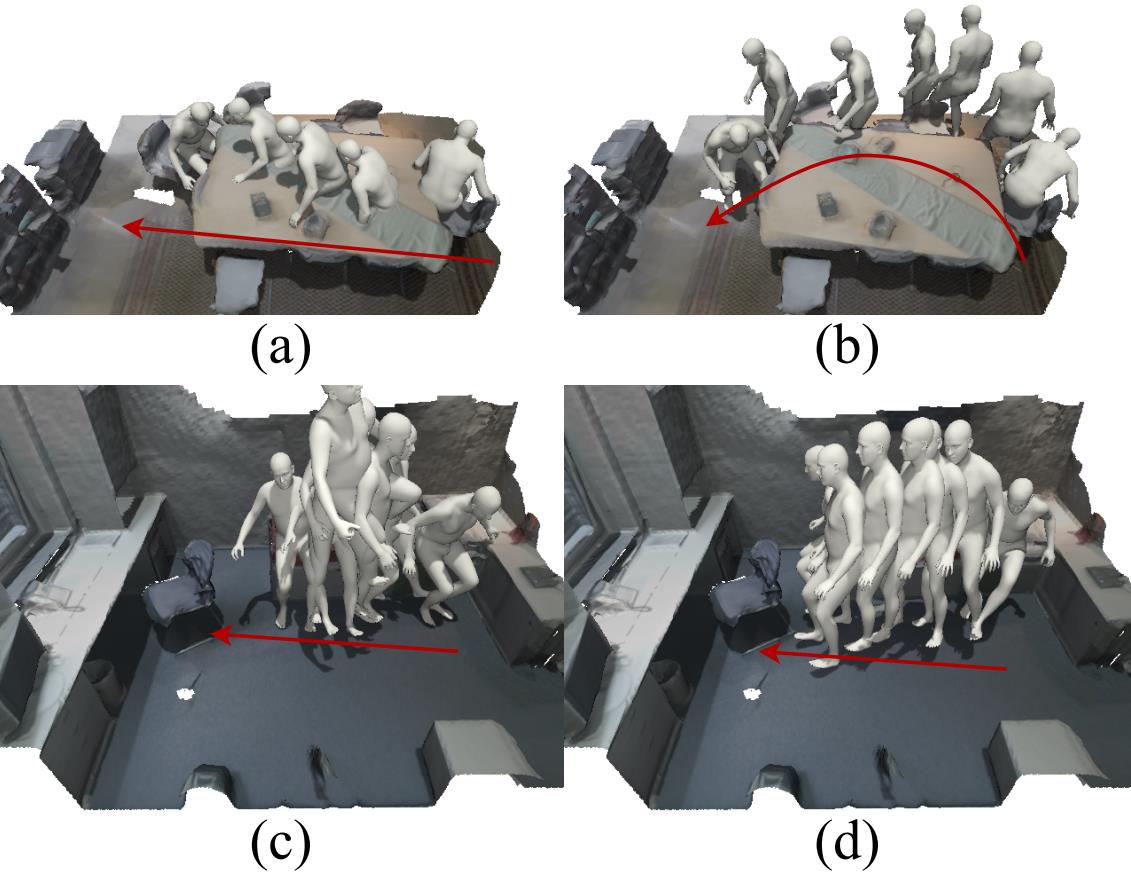}
	\small\caption{\textbf{Effect of the planned path.} (a) is the result without planning module and (b) is based on planning module. (c) is the results for only using translations of planned path as the position encoding for our Path Refiner in Section 3.4 and (d) is result from our method.}
	\label{fig:motion_ablation}
	\vspace{-0.3cm}
\end{figure}
\noindent\paragraph{Planned Path for Motion Synthesis.}
As shown in Figure~\ref{fig:motion_ablation}, we show the effectiveness of using planned paths in the procedure of motion synthesis. 
Firstly, without the additional positional encoding from the planned path, the synthesized motion can not follow the planned path and penetrate to the table, as shown in Figure~\ref{fig:motion_ablation} (a). 
Besides, we find that both the translation and orientation for the planned path are also crucial for motion synthesis. 
As shown in Figure~\ref{fig:motion_ablation} (c), the Path Refiner synthesizes unnatural orientations for human motions without encoding the orientation of the planned path into the positional encoding as Section 3.4. 
Instead, as shown in Figure~\ref{fig:motion_ablation} (b) and (d), our method can synthesize natural human motion with the translation and orientation of planned path as the additional positional encoding for our Path Refiner.

\section{Acknowledgement}
This study is supported under the General Research Fund (GRF) of Hong Kong (No.,14205719), the RIE2020 Industry Alignment Fund–Industry Collaboration Projects (IAF-ICP) Funding Initiative, as well as cash and in-kind contribution from the industry partner(s).

{\small
\bibliographystyle{ieee_fullname}
\bibliography{egbib}

\begin{thebibliography}{10}\itemsep=-1pt

\bibitem{barsoum2018hp}
Emad Barsoum, John Kender, and Zicheng Liu.
\newblock Hp-gan: Probabilistic 3d human motion prediction via gan.
\newblock In {\em Proceedings of the IEEE conference on computer vision and
  pattern recognition workshops}, 2018.

\bibitem{cai2018deep}
Haoye Cai, Chunyan Bai, Yu-Wing Tai, and Chi-Keung Tang.
\newblock Deep video generation, prediction and completion of human action
  sequences.
\newblock In {\em Proceedings of the European Conference on Computer Vision
  (ECCV)}, 2018.

\bibitem{cai2021unified}
Yujun Cai, Yiwei Wang, Yiheng Zhu, Tat-Jen Cham, Jianfei Cai, Junsong Yuan, Jun
  Liu, Chuanxia Zheng, Sijie Yan, Henghui Ding, et~al.
\newblock A unified 3d human motion synthesis model via conditional variational
  auto-encoder.
\newblock In {\em International Conference on Computer Vision}, 2021.

\bibitem{cao2020long}
Zhe Cao, Hang Gao, Karttikeya Mangalam, Qi-Zhi Cai, Minh Vo, and Jitendra
  Malik.
\newblock Long-term human motion prediction with scene context.
\newblock In {\em ECCV}, 2020.

\bibitem{carpin2006randomized}
Stefano Carpin.
\newblock Randomized motion planning: a tutorial.
\newblock {\em International Journal of Robotics and Automation}, 2006.

\bibitem{Matterport3D}
Angel Chang, Angela Dai, Thomas Funkhouser, Maciej Halber, Matthias Niessner,
  Manolis Savva, Shuran Song, Andy Zeng, and Yinda Zhang.
\newblock Matterport3d: Learning from rgb-d data in indoor environments.
\newblock In {\em International Conference on 3D Vision (3DV)}, 2017.

\bibitem{Cui_2020_CVPR}
Qiongjie Cui, Huaijiang Sun, and Fei Yang.
\newblock Learning dynamic relationships for 3d human motion prediction.
\newblock In {\em Proceedings of the IEEE/CVF Conference on Computer Vision and
  Pattern Recognition (CVPR)}, June 2020.

\bibitem{fragkiadaki2015recurrent}
Katerina Fragkiadaki, Sergey Levine, Panna Felsen, and Jitendra Malik.
\newblock Recurrent network models for human dynamics.
\newblock In {\em Proceedings of the IEEE International Conference on Computer
  Vision}, 2015.

\bibitem{chuan2020action2motion}
Chuan Guo, Xinxin Zuo, Sen Wang, Shihao Zou, Qingyao Sun, Annan Deng, Minglun
  Gong, and Li Cheng.
\newblock Action2motion: Conditioned generation of 3d human motions.
\newblock In {\em Proceedings of the 28th ACM International Conference on
  Multimedia (MM '20)}, 2020.

\bibitem{hart1968formal}
Peter~E Hart, Nils~J Nilsson, and Bertram Raphael.
\newblock A formal basis for the heuristic determination of minimum cost paths.
\newblock {\em IEEE transactions on Systems Science and Cybernetics}, 1968.

\bibitem{harvey2020robust}
F{\'e}lix~G Harvey, Mike Yurick, Derek Nowrouzezahrai, and Christopher Pal.
\newblock Robust motion in-betweening.
\newblock {\em ACM Transactions on Graphics (TOG)}, 2020.

\bibitem{hassan_samp_2021}
Mohamed Hassan, Duygu Ceylan, Ruben Villegas, Jun Saito, Jimei Yang, Yi Zhou,
  and Michael Black.
\newblock Stochastic scene-aware motion prediction.
\newblock In {\em Proceedings of the International Conference on Computer
  Vision 2021}, Oct. 2021.

\bibitem{PROX:2019}
Mohamed Hassan, Vasileios Choutas, Dimitrios Tzionas, and Michael~J. Black.
\newblock Resolving {3D} human pose ambiguities with {3D} scene constraints.
\newblock In {\em International Conference on Computer Vision}, 2019.

\bibitem{Hassan:CVPR:2021}
Mohamed Hassan, Partha Ghosh, Joachim Tesch, Dimitrios Tzionas, and Michael~J.
  Black.
\newblock Populating {3D} scenes by learning human-scene interaction.
\newblock In {\em Conference Computer Vision and Pattern Recognition}, 2021.

\bibitem{hochreiter1997long}
Sepp Hochreiter and J{\"u}rgen Schmidhuber.
\newblock Long short-term memory.
\newblock {\em Neural computation}, 1997.

\bibitem{kingma2014adam}
Diederik~P Kingma and Jimmy Ba.
\newblock Adam: A method for stochastic optimization.
\newblock 2014.

\bibitem{kingma2013auto}
Diederik~P Kingma and Max Welling.
\newblock Auto-encoding variational bayes.
\newblock {\em arXiv preprint arXiv:1312.6114}, 2013.

\bibitem{KipfW17}
Thomas~N. Kipf and Max Welling.
\newblock Semi-supervised classification with graph convolutional networks.
\newblock In {\em International Conference on Learning Representations}, 2017.

\bibitem{Li_2020_CVPR}
Maosen Li, Siheng Chen, Yangheng Zhao, Ya Zhang, Yanfeng Wang, and Qi Tian.
\newblock Dynamic multiscale graph neural networks for 3d skeleton based human
  motion prediction.
\newblock In {\em Proceedings of the IEEE/CVF Conference on Computer Vision and
  Pattern Recognition (CVPR)}, 2020.

\bibitem{li2017auto}
Zimo Li, Yi Zhou, Shuangjiu Xiao, Chong He, Zeng Huang, and Hao Li.
\newblock Auto-conditioned recurrent networks for extended complex human motion
  synthesis.
\newblock {\em arXiv preprint arXiv:1707.05363}, 2017.

\bibitem{LiuN89}
Dong~C. Liu and Jorge Nocedal.
\newblock On the limited memory {BFGS} method for large scale optimization.
\newblock {\em Mathematical Programming}, 1989.

\bibitem{mao2019learning}
Wei Mao, Miaomiao Liu, Mathieu Salzmann, and Hongdong Li.
\newblock Learning trajectory dependencies for human motion prediction.
\newblock In {\em Proceedings of the IEEE International Conference on Computer
  Vision}, 2019.

\bibitem{martinez2017human}
Julieta Martinez, Michael~J Black, and Javier Romero.
\newblock On human motion prediction using recurrent neural networks.
\newblock In {\em Proceedings of the IEEE Conference on Computer Vision and
  Pattern Recognition}, 2017.

\bibitem{pavlakos2019expressive}
Georgios Pavlakos, Vasileios Choutas, Nima Ghorbani, Timo Bolkart, Ahmed~AA
  Osman, Dimitrios Tzionas, and Michael~J Black.
\newblock Expressive body capture: 3d hands, face, and body from a single
  image.
\newblock In {\em Conference on Computer Vision and Pattern Recognition}, 2019.

\bibitem{pavllo2018quaternet}
Dario Pavllo, David Grangier, and Michael Auli.
\newblock Quaternet: A quaternion-based recurrent model for human motion.
\newblock {\em arXiv preprint arXiv:1805.06485}, 2018.

\bibitem{petrovich21actor}
Mathis Petrovich, Michael~J. Black, and G{\"u}l Varol.
\newblock Action-conditioned 3{D} human motion synthesis with transformer
  {VAE}.
\newblock In {\em International Conference on Computer Vision (ICCV)}, 2021.

\bibitem{qi2017pointnet}
Charles~R Qi, Hao Su, Kaichun Mo, and Leonidas~J Guibas.
\newblock Pointnet: Deep learning on point sets for 3d classification and
  segmentation.
\newblock In {\em Proceedings of the IEEE conference on computer vision and
  pattern recognition}, 2017.

\bibitem{sohn2015learning}
Kihyuk Sohn, Honglak Lee, and Xinchen Yan.
\newblock Learning structured output representation using deep conditional
  generative models.
\newblock {\em Advances in neural information processing systems}, 2015.

\bibitem{StarkeZKS19}
Sebastian Starke, He Zhang, Taku Komura, and Jun Saito.
\newblock Neural state machine for character-scene interactions.
\newblock {\em {ACM} Transactions on Graphics.}, 2019.

\bibitem{sutskever2014sequence}
Ilya Sutskever, Oriol Vinyals, and Quoc~V Le.
\newblock Sequence to sequence learning with neural networks.
\newblock In {\em Advances in neural information processing systems}, 2014.

\bibitem{vsvestka1998probabilistic}
Petr {\v{S}}vestka and Markus~Hendrik Overmars.
\newblock Probabilistic path planning.
\newblock 1998.

\bibitem{GRAB:2020}
Omid Taheri, Nima Ghorbani, Michael~J. Black, and Dimitrios Tzionas.
\newblock {GRAB}: A dataset of whole-body human grasping of objects.
\newblock In {\em European Conference on Computer Vision}, 2020.

\bibitem{vaswani2017attention}
Ashish Vaswani, Noam Shazeer, Niki Parmar, Jakob Uszkoreit, Llion Jones,
  Aidan~N Gomez, {\L}ukasz Kaiser, and Illia Polosukhin.
\newblock Attention is all you need.
\newblock In {\em Advances in neural information processing systems}, 2017.

\bibitem{wang2021synthesizing}
Jiashun Wang, Huazhe Xu, Jingwei Xu, Sifei Liu, and Xiaolong Wang.
\newblock Synthesizing long-term 3d human motion and interaction in 3d scenes.
\newblock In {\em Proceedings of the IEEE/CVF Conference on Computer Vision and
  Pattern Recognition}, 2021.

\bibitem{Wang_2021_CVPR}
Jingbo Wang, Sijie Yan, Bo Dai, and Dahua Lin.
\newblock Scene-aware generative network for human motion synthesis.
\newblock In {\em Conference on Computer Vision and Pattern Recognition}, 2021.

\bibitem{xu2020hierarchical}
Jingwei Xu, Huazhe Xu, Bingbing Ni, Xiaokang Yang, Xiaolong Wang, and Trevor
  Darrell.
\newblock Hierarchical style-based networks for motion synthesis.
\newblock In {\em European Conference on Computer Vision}, 2020.

\bibitem{yan2019convolutional}
Sijie Yan, Zhizhong Li, Yuanjun Xiong, Huahan Yan, and Dahua Lin.
\newblock Convolutional sequence generation for skeleton-based action
  synthesis.
\newblock In {\em Proceedings of the IEEE International Conference on Computer
  Vision}, 2019.

\bibitem{yan2018spatial}
Sijie Yan, Yuanjun Xiong, and Dahua Lin.
\newblock Spatial temporal graph convolutional networks for skeleton-based
  action recognition.
\newblock In {\em AAAI}, 2018.

\bibitem{yang2018pose}
Ceyuan Yang, Zhe Wang, Xinge Zhu, Chen Huang, Jianping Shi, and Dahua Lin.
\newblock Pose guided human video generation.
\newblock In {\em Proceedings of the European Conference on Computer Vision
  (ECCV)}, 2018.

\bibitem{yuan2020dlow}
Ye Yuan and Kris Kitani.
\newblock Dlow: Diversifying latent flows for diverse human motion prediction.
\newblock In {\em Proceedings of the European Conference on Computer Vision
  (ECCV)}, 2020.

\bibitem{PLACE:3DV:2020}
Siwei Zhang, Yan Zhang, Qianli Ma, Michael~J. Black, and Siyu Tang.
\newblock {PLACE}: Proximity learning of articulation and contact in {3D}
  environments.
\newblock In {\em International Conference on 3D Vision (3DV)}, Nov. 2020.

\bibitem{zhang2021we}
Yan Zhang, Michael~J Black, and Siyu Tang.
\newblock We are more than our joints: Predicting how 3d bodies move.
\newblock In {\em Conference on Computer Vision and Pattern Recognition}, 2021.

\bibitem{PSI:2019}
Yan Zhang, Mohamed Hassan, Heiko Neumann, Michael~J. Black, and Siyu Tang.
\newblock Generating 3d people in scenes without people.
\newblock In {\em Computer Vision and Pattern Recognition (CVPR)}, 2020.

\bibitem{zhou2019continuity}
Yi Zhou, Connelly Barnes, Jingwan Lu, Jimei Yang, and Hao Li.
\newblock On the continuity of rotation representations in neural networks.
\newblock In {\em Proceedings of the IEEE/CVF Conference on Computer Vision and
  Pattern Recognition (CVPR)}, 2019.

\end{thebibliography}
}

\end{document}